\documentclass[runningheads]{llncs}
\usepackage{graphicx,subcaption}
\usepackage{amsmath}
\usepackage{amsfonts}
\usepackage{booktabs}
\usepackage{todonotes}
\usepackage{semantic}
\usepackage{multirow}
\usepackage{array}
\usepackage[section]{placeins}
\usepackage[square,sort,comma,numbers]{natbib}
\usepackage{algpseudocode}
\usepackage{algorithm}
\usepackage{hyperref}

\newcolumntype{C}[1]{>{\centering\let\newline\\\arraybackslash\hspace{0pt}}m{#1}}
\newcommand{\el}{$\mathcal{EL}^{++}$}

\newcommand{\interp}[1]{{#1}^\mathcal{I}}

\begin{document}
\title{Title}
\title{Enhancing Geometric Ontology Embeddings for $\mathcal{EL}^{++}$
  with Negative Sampling and Deductive Closure Filtering}
\titlerunning{Negative Sampling and Deductive Closure Filtering}
\author{Olga Mashkova\inst{1}\orcidID{0000-0002-4916-1660} \and
Fernando Zhapa-Camacho\inst{1}\orcidID{0000-0002-0710-2259} \and
Robert Hoehndorf\inst{1}\orcidID{0000-0001-8149-5890}}
\authorrunning{O.Mashkova et al.}
\institute{Computer Science Program, Computer, Electrical, and
  Mathematical Sciences \& Engineering Division, King Abdullah
  University of Science and Technology, Thuwal 23955, Saudi Arabia \\
\email{\{olga.mashkova,fernando.zhapacamacho,robert.hoehndorf\}@kaust.edu.sa}}
\maketitle              
\begin{abstract}
Ontology embeddings map classes, relations, and individuals in
ontologies into $\mathbb{R}^n$, and within $\mathbb{R}^n$ similarity
between entities can be computed or new axioms inferred.
For ontologies in the Description Logic $\mathcal{EL}^{++}$, several
embedding methods have been developed that explicitly generate models
of an ontology.
However, these methods suffer from some limitations; 
they do not distinguish between statements that are unprovable and
provably false, and therefore they may use entailed statements as
negatives. Furthermore, they do not utilize the deductive closure of
an ontology to identify statements that are inferred but not asserted.
We evaluated a set of embedding methods for $\mathcal{EL}^{++}$
ontologies based on high-dimensional ball representation of concept
descriptions, incorporating several modifications that aim to make use
of the ontology deductive closure.
In particular, we designed novel negative losses that account both for
the deductive closure and different types of negatives. 
We demonstrate that our embedding methods improve over the baseline
ontology embedding in the task of knowledge base or ontology
completion.

\end{abstract}
\section{Introduction}

Several methods have been developed to embed Description Logic
theories or ontologies in vector spaces~\cite{owl2vecstar,dl2vec,kulmanov2019embeddings,mondal2021emel++,peng2022description,xiong2022faithful,jackermeier2023box,ozcep2023embedding}. These embedding
methods preserve some aspects of the semantics in the vector space,
and may enable the computation of semantic similarity, inferring
axioms that are entailed, and predicting axioms that are not entailed
but may be added to the theory. For the lightweight Description Logic
\el, several geometric embedding methods have been
developed~\cite{kulmanov2019embeddings,mondal2021emel++,xiong2022faithful,jackermeier2023box,ozcep2023embedding}. They
can be proven to ``faithfully'' approximate a model in the sense that,
if a certain optimization objective is reached (usually, a loss
function reduced to $0$), the embedding method has constructed a model
of the \el theory. Geometric model construction enables the execution
of various tasks. These tasks include knowledge base completion and
subsumption prediction via either testing the truth of a statement
under consideration in a single (approximate) model or aggregating
truth values over multiple models.

Advances on different geometric embedding methods
have usually focused on the expressiveness of the embedding methods;
originally, hyperballs\cite{kulmanov2019embeddings} where used to represent the interpretation of
concept symbols, yet hyperballs are not closed under
intersection. Therefore, axis-aligned boxes were
introduced~\cite{peng2022description,xiong2022faithful,jackermeier2023box}. Furthermore, \el allows for axioms pertaining to relations, and several methods have extended the way in which
relations are modeled~\cite{jackermeier2023box,kulmanov2019embeddings,xiong2022faithful}. However, there are several aspects of
geometric embeddings that have not yet been investigated. In
particular, for \el, there are sound and complete reasoners with
efficient implementations that scale to very large knowledge
bases~\cite{elk}; it may therefore be possible to utilize a deductive reasoner
together with the embedding process to improve generation of
embeddings that represent geometric models.

We evaluate geometric embedding methods and incorporate deductive
inference into the training process. We use the
{\it ELEmbeddings}~\cite{kulmanov2019embeddings} model for our experiments
due to its simplicity; however, our results also apply to other
geometric embedding methods for \el.

Our main contributions are as follows:

\begin{itemize}
\item We investigate and reveal biases in some evaluation datasets
  that are related to how the task of knowledge base completion is
  formulated, and demonstrate that, due to these biases, even when
  models collapse, predictive performance can be high.
\item We introduce loss functions that avoid zero gradients and
  improve the task of knowledge base completion.
\item We introduce a fast approximate algorithm for computing the
  deductive closure of an \el theory and use it to improve negative
  sampling during model training.
\item We propose loss functions that incorporate negative samples in
  most normal forms.
\end{itemize}

\section{Preliminaries}

\subsection{Description Logic \el} \label{dl}

Let $\Sigma = (\mathbf{C}, \mathbf{R}, \mathbf{I})$ be a signature
with set $\mathbf{C}$ of concept names, $\mathbf{R}$ of role names,
and $\mathbf{I}$ of individual names. Given $A, B \in \mathbf{C}$,
$r \in \mathbf{R}$, and $a, b \in \mathbf{I}$, \el concept
descriptions are constructed with the grammar
$ \bot \mid \top \mid A \sqcap B \mid \exists r.A \mid \{a\}$. ABox
axioms are of the form $A(a)$ and $r(a,b)$, TBox axioms are of the
form $A \sqsubseteq B$, and RBox axioms are of the form
$r_1 \circ r_2 \circ \dots \circ r_n \sqsubseteq r$. \el
\emph{generalized concept inclusions} (GCIs)
and \emph{role inclusions} (RIs)
can be normalized to follow one of these forms~\cite{baader2005pushing}:
$C \sqsubseteq D$
(GCI0), $C \sqcap D \sqsubseteq E$ (GCI1), $C \sqsubseteq \exists R.D$
(GCI2), $\exists R.C \sqsubseteq D$ (GCI3), $C \sqsubseteq \bot$
(GCI0-BOT), $C \sqcap D \sqsubseteq \bot$ (GCI1-BOT),
$\exists R.C \sqsubseteq \bot$ (GCI3-BOT) and $r \sqsubseteq s$ (RI0),
$r_1 \circ r_2 \sqsubseteq s$ (RI1), respectively.

To define the semantics of an $\mathcal{EL}^{++}$ theory, we
use~\cite{baader2005pushing}
an \emph{interpretation domain} $\Delta^{\mathcal{I}}$ and an
\emph{interpretation function} $\cdot^{\mathcal{I}}$.  For every
concept $A \in \mathbf{C}$, $\interp{A} \subseteq \interp{\Delta}$;
individual $a \in \mathbf{I}$, $\interp{a} \in \interp{\Delta}$; role
$r \in \mathbf{R}$,
$\interp{r} \in \interp{\Delta} \times \interp{\Delta}$. Furthermore,
the semantics for other $\mathcal{EL}^{++}$ constructs are the
following (omitting concrete domains and role inclusions):
\begin{equation}
  \nonumber
  \begin{split}
    {\bot}^{\mathcal{I}} &= \emptyset\\
    {\top}^{\mathcal{I}} &=\Delta^{\mathcal{I}},\\
    (A \sqcap B)^{\mathcal{I}} &= A^{\mathcal{I}} \cap B^{\mathcal{I}}, \\
    (\exists r . A)^{\mathcal{I}} &=\left\{a \in \Delta^{\mathcal{I}}
    \mid \exists \thinspace b : ((a, b) \in r^{\mathcal{I}} \land b \in \interp{A})\right\},\\
   \interp{({a})} &= \{a\}
  \end{split}
  \quad \quad
  \begin{split}
   \end{split}
\end{equation}

An interpretation $\mathcal{I}$ is a model for an axiom
${C} \sqsubseteq {D}$ if and only if
$\interp{C} \subseteq \interp{D}$, for an axiom $B(a)$ if and only if
$\interp{a} \in \interp{B}$; and for an axiom $r(a,b)$ if and only if
$(\interp{a},\interp{b}) \in \interp{r}$~\cite{dl_handbook}.

\subsection{Knowledge Base Completion}

The task of knowledge base completion is the addition (or prediction)
of axioms to a knowledge base that are not explicitly represented. We
call the task ``ontology completion'' when exclusively TBox axioms are
predicted. The task of knowledge base completion may encompass both
deductive~\cite{sato2018deductive,jiang2012combining} and
inductive~\cite{bouraoui2017inductive,d2012ontology} inference
processes and give rise to two subtly different tasks: adding only
``novel'' axioms to a knowledge base that are {\em not} in the
deductive closure of the knowledge base, and adding axioms that are in
the deductive closure as well as some ``novel'' axioms that are not
deductively inferred; both tasks are related but differ in how they
are evaluated.
  
Inductive inference, analogously to knowledge graph
completion~\cite{chen2020knowledge}, predicts axioms based on patterns and regularities
within the knowledge base.  Knowledge base completion, or ontology
completion, can be further distinguished based on the information that
is used to predict ``novel'' axioms.  We distinguish between two
approaches to knowledge base completion: (1) knowledge base completion
which relies solely on (formalized) information within the knowledge
base to predict new axioms, and (2) knowledge base completion which
incorporates side information, such as text, to enhance the prediction
of new axioms. Here, we mainly consider the first case.

\section{Related Work}

\subsection{Graph-Based Ontology Embeddings }
Graph-based ontology embeddings rely on a construction (projection) of
graphs from ontology axioms mapping ontology classes, individuals and
roles to nodes and labeled edges~\cite{graph_projections}. Embeddings
for nodes and edge labels are optimized using Knowledge Graph
Embedding (KGE) methods~\cite{wang2017knowledge}.
These type of methods have been shown effective on knowledge
base and ontology completion~\cite{owl2vecstar} and have been applied
to domain-specific tasks such as protein--protein interaction
prediction~\cite{owl2vecstar} or gene--disease association
prediction~\cite{dl2vec}.  Graph-based methods rely on adjacency
information of the ontology structure but cannot easily handle logical
operators and do not 
approximate ontology models.  Therefore, graph-based
methods are not ``faithful'', i.e., do not approximate models, do not
allow determining whether statements are ``true'' in these models, and
therefore cannot be used to
perform semantic entailment.

\subsection{Geometric-Based Ontology Embeddings}

Multiple methods have been developed for the geometric construction of
models for the $\mathcal{EL}^{++}$ language.
ELEmbeddings~\cite{kulmanov2019embeddings} constructs an
interpretation of concept names as sets of points lying within an open
$n$-dimensional ball and generates an interpretation of role names as
the set of pairs of points that are separated by a vector in
$\mathbb{R}^n$, i.e., by the embedding of the role name.
EmEL++~\cite{mondal2021emel++} extends ELEmbeddings with more
expressive constructs such as role chains and role
inclusions. ELBE~\cite{peng2022description} and
BoxEL~\cite{xiong2022faithful} use $n$-dimensional axis-aligned boxes
to represent concepts, which has an advantage over balls because the
intersection of two axis-aligned boxes is a box whereas the
intersection of two $n$-balls is not an $n$-ball. BoxEL additionally
preserves ABox facilitating a more accurate representation of
knowledge base's logical structure by ensuring, e.g., that an entity
has the minimal volume.  Box$^2$EL~\cite{jackermeier2023box}
represents ontology roles more expressively with two boxes encoding
the semantics of the domain and codomain of roles. Box$^2$EL enables
the expression of one-to-many relations as opposed to other
methods. Axis-aligned cone-shaped geometric model introduced
in~\cite{ozcep2023embedding} deals with $\mathcal{ALC}$ ontologies and
allows for full negation of concepts and existential quantification by
construction of convex sets in ${\mathbb R}^n$.  This work has not yet
been implemented or evaluated in an application.

\subsection{Knowledge Base Completion Task}

Several recent advancements in the knowledge base completion rely on
side information as included in Large Language Models (LLMs).
\cite{ji2023ontology} explores how pretrained language models can be
utilized for incorporating one ontology into another, with the main
focus on inconsistency handling and ontology
coherence. HalTon~\cite{cao2023event} addresses the task of event
ontology completion via simultaneous event clustering, hierarchy
expansion and type naming utilizing BERT~\cite{devlin2018bert} for
instance encoding. \cite{li2024ontology} formulates knowledge base
completion task as a Natural Language Inference (NLI) problem and
examines how this approach may be combined with concept embeddings for
identifying missing knowledge in ontologies. As for other approaches,
\cite{mevznar2022ontology} proposes a method that converts an ontology
into a graph to recommend missing edges using structure-only link
analysis methods, \cite{shiraishi2024self} constructs matrix-based
ontology embeddings which capture the global and local information for
subsumption prediction. All these methods use side information from
LLMs and would not be applicable, for example, in the case where a
knowledge base is private or consists of only identifiers; we do not
consider methods based on pre-trained LLMs here as baselines.

\section{Methods}

\subsection{Datasets}

Following previous
works~\cite{kulmanov2019embeddings,peng2022description,jackermeier2023box}
we use common benchmarks for the prediction of protein--protein
interactions (PPIs). We also reorganize the same data for the task of
protein function prediction.  For our experiments we use four
datasets; each of them consists of the Gene
Ontology~\cite{gene2015gene} with all its axioms, protein--protein
interactions (PPIs) and protein function axioms extracted from the
STRING database~\cite{mering2003string}; we use one dataset focusing
on only yeast and another dataset focusing on only human proteins.  GO
is formalized using OWL 2 EL~\cite{horrocksobo}.

For PPI yeast network we use the built-in dataset {\tt
  PPIYeastDataset} available in the
mOWL~\cite{10.1093/bioinformatics/btac811} Python library (release
0.2.1) where axioms of interest are split randomly into train,
validation and test datasets in ratio 90:5:5 keeping pairs of
symmetric PPI axioms within the same dataset, and other axioms are
placed into the training part; validation and test sets are made up of
TBox axioms of type
$\{P_1\} \sqsubseteq \exists interacts\_with.\{P_2\}$ where $P_1, P_2$
are protein names.  In case of yeast proteins, the GO version released
on 2021-10-20 and the STRING database version 11.5 were
used. Alongside with the yeast $interacts\_with$ dataset we collected
the yeast $has\_function$ dataset organized in the same manner with
validation and test parts containing TBox axioms of type
$\{P\} \sqsubseteq \exists has\_function.\{GO\}$. The human
$interacts\_with$ and $has\_function$ datasets were built from STRING
PPI human network (version 10.5) and GO released on 2018-12-28.  Based
on the information in the STRING database, in PPI yeast, the {\em
  interacts\_with} relation is symmetric and the dataset is closed
against symmetric interactions; the PPI human dataset does not always
contain the inverse of interactions and is not closed against
symmetry.  We normalize all ontology axioms using the implementation
of the jcel~\cite{mendez2012jcel} reasoner, accessed through the mOWL
library~\cite{10.1093/bioinformatics/btac811}.  Role inclusion axioms
are ignored since we experiment with modifications of the {\it ELEmbeddings} method where role inclusion axioms are omitted as well. The number of GCIs of each type in the datasets can be found in the Appendix~\ref{app:gci_numbers}.

\subsection{Objective Functions}
{\it ELEmbeddings} use a single loss for ``negatives'', i.e., axioms
that are not included in the knowledge base; the loss is used only for
axioms of the form $C \sqsubseteq \exists R.D$ which are randomly
sampled, and negatives are not considered for other normal forms. We
add three more ``negative'' losses:
$C \sqsubseteq D$, $C \sqcap D \sqsubseteq E$, and $\exists R.C \sqsubseteq D$:

\begin{align} \label{eq1}
\begin{split}
    loss_{C \not\sqsubseteq D}(c, d) = \\ 
    = l(r_{\eta}(c) + r_{\eta}(d) - \|f_{\eta}(c) - f_{\eta}(d))\| + \gamma) + \\
    + | \|f_{\eta}(c)\| - 1| + | \|f_{\eta}(d)\| - 1|
\end{split}
\end{align}

\begin{align} \label{eq2}
\begin{split}
    loss_{C \sqcap D \not\sqsubseteq E}(c, d, e) = \\ 
    = l(- r_{\eta}(c) - r_{\eta}(d) + \|f_{\eta}(c) - f_{\eta}(d))\| - \gamma) + \\
    + l(r_{\eta}(c) - \|f_{\eta}(c) - f_{\eta}(e))\| + \gamma) + \\
    + l(r_{\eta}(d) - \|f_{\eta}(d) - f_{\eta}(e))\| + \gamma) + \\
    + | \|f_{\eta}(c)\| - 1| + | \|f_{\eta}(d)\| - 1| + | \|f_{\eta}(e)\| - 1|
\end{split}
\end{align}

\begin{align} \label{eq3}
\begin{split}
    loss_{\exists R.C\not\sqsubseteq D}(r, c, d) = \\ 
    = l(r_{\eta}(c) + r_{\eta}(d) - \|f_{\eta}(c) - f_{\eta}(r) - f_{\eta}(d))\| + \gamma) + \\
    + | \|f_{\eta}(c)\| - 1| + | \|f_{\eta}(d)\| - 1|
\end{split}
\end{align}
Here, $l$ denotes a function that determines the behavior of the loss
when the axiom is true (for positive cases) or not true (for negative
cases); in our case, we consider ReLU and LeakyReLU; $\gamma$ stands
for a margin parameter. We employ notations from the {\it
  ELEmbeddings} method: $r_{\eta}(c), \thinspace r_{\eta}(d),
\thinspace r_{\eta}(e)$ and $f_{\eta}(c), \thinspace f_{\eta}(d),
\thinspace f_{\eta}(e)$ denote the radius and the ball center
associated with classes $c, d, e$, respectively, $f_{\eta}(r)$ denotes
the embedding vector associated with relation $r$. There is a geometrical part as well as a
regularization part for each new negative loss forcing class centers
to lie on a unit $\ell_2-$sphere. 
Negative loss~\ref{eq3} is constructed similarly to $C \sqcap D \sqsubseteq E$ loss: the first part penalizes non-overlap of $C$ and $D$ classes (we do not consider disjointness case since for every class $X$ we have $\bot \sqsubseteq X$); the second and the third part force the center corresponding to $E$ not to lie in the intersection of balls associated with $C$ and $D$. Here we do not consider constraints on radius of the ball for $E$ class and focus only on relative positions of $C, D$ and $E$ class centers and overlapping of $n$-balls representing $C$ and $D$.
In our experiments, we also use a
relaxed regularization where $\|f_{\eta}(c)\| = R$ is replaced with
$\|f_{\eta}(c)\| \leq R$ on $n$-ball centers representing concepts
forcing them to lie inside the corresponding closed ball of radius $R$
centered at $0$. Relaxed version of regularization may allow for more
accurate representation of a knowledge base since it is not forcing
all ball centers corresponding to concept names to lie on a unit
sphere. 

\subsection{Deductive Closure: Negatives Filtration}

The {\it deductive closure} of a theory $T$ refers to the smallest set
containing all statements which can be inferred by deductive reasoning
over $T$; for a given deductive relation $\vdash$, we call
$T^\vdash = \{ \phi \thinspace | \thinspace T \vdash \phi \}$ the
deductive closure of $T$.  In knowledge bases, the deductive closure
is usually not identical to the asserted axioms in the knowledge base,
and will contain axioms that are non-trivial; it is also usually
infinite.

Representing the deductive closure is challenging since it is
infinite, but in \el any knowledge base can be normalized to one of
the seven normal forms; therefore, we can compute the deductive
closure with respect to these normal forms. However, existing \el
reasoners such as ELK~\cite{elk} compute all axioms of the form
$C \sqsubseteq D$ in the deductive closure but not the other normal
forms. We use the inferences computed by ELK (of the form
$C\sqsubseteq D$) to design an algorithm that computes the deductive
closure with respect to the \el normal forms; the algorithm implements
sound but incomplete inference rules (see Algorithm 1 for further
details); specifically, it computes entailed axioms for all normal
forms based on the concept hierarchy pre-computed by ELK.

\subsection{Training Procedure}~\label{training_procedure}
To address the issue of data imbalance (see
Appendix~\ref{app:gci_numbers}), i.e., the imbalance between the
number of axioms of different normal forms represented in a knowledge
base which may have an impact on how well certain types of axioms are
represented in the embedding space, we weigh individual GCI losses
based on frequency of the axiom types sampled during one epoch. All
models are optimized with respect to the weighted sum of individual GCI
losses (here we define the loss in most general case using all
positive and all negative losses):
\begin{align}
\begin{split}
    {\mathcal L} = w_{C \sqsubseteq D} \cdot l_{C \sqsubseteq D} + w_{C \sqcap D \sqsubseteq E} \cdot l_{C \sqcap D \sqsubseteq E} + w_{C \sqsubseteq \exists R.D} \cdot l_{C \sqsubseteq \exists R.D} + \\ 
    + w_{\exists R.C \sqsubseteq D} \cdot l_{\exists R.C \sqsubseteq D} + w_{C \sqsubseteq \bot} \cdot l_{C \sqsubseteq \bot} + w_{\exists R.C \sqsubseteq \bot} \cdot l_{\exists R.C \sqsubseteq \bot} + \\
    + w_{C \not\sqsubseteq D} \cdot l_{C \not\sqsubseteq D} + w_{C \sqcap D \not\sqsubseteq E} \cdot l_{C \sqcap D \not\sqsubseteq E} + w_{C \not\sqsubseteq \exists R.D} \cdot l_{C \not\sqsubseteq \exists R.D} + \\
    + w_{\exists R.C \not\sqsubseteq D} \cdot l_{\exists R.C
  \not\sqsubseteq D}
\end{split}
\end{align}
To study the phenomenon of biases in data affecting model training and
performance, we 
build a `naive' model which predicts only based on the frequency with
which a class appears in an axiom. Intuitively, it is designed to
resemble predictions based on node degree in knowledge graphs:
\begin{equation}
  score_{C \sqsubseteq \exists R.D}(c, r, d) = \frac{\sum_{c'} M_r(c', d)}{\sum_{k, l} M_r(k, l)}
\end{equation}

All model architectures are built using
mOWL~\cite{10.1093/bioinformatics/btac811} library on top of mOWL's
base models. All models were trained using the same fixed random
seed. Training code for all experiments and models is available on
\url{https://github.com/bio-ontology-research-group/geometric_embeddings}.

All models are trained for 400 epochs with batch size of
32,768. Training and optimization is performed using Pytorch with Adam
optimizer~\cite{kingma2014adam} and ReduceLROnPlateau scheduler with
patience parameter $10$. We apply early stopping if validation loss does
not improve for $20$ epochs. Hyperparameters are tuned using
grid search over the following set: margin
$\gamma \in \{-0.1, -0.01, 0, 0.01, 0.1\}$, embedding dimension $\{50,
100, 200, 400\}$, regularization radius $R \in \{1, 2\}$, learning rate
$\{0.01, 0.001, 0.0001\}$. For {\it ELEmbeddings}, the strict
version of regularization $\|f_{\eta}(c)\| = R$ was used with $R =
1$; see Appendix~\ref{sec:hyper} for details on optimal
hyperparameters used.

\subsection{Evaluation score and metrics}
We predict GCI2 axioms of type
$\{P_1\} \sqsubseteq \exists interacts\_with.\{P_2\}$ or
$\{P\} \sqsubseteq \\ \sqsubseteq \exists has\_function.\{GO\}$
depending on the dataset. As the core evaluation score we use the
scoring function introduced in {\it ELEmbeddings}:
\begin{align}
\begin{split}
  score_{C \sqsubseteq \exists R.D}(c, r, d) = \\ 
  = -l(- r_{\eta}(c) - r_{\eta}(d) + \|f_{\eta}(c) + f_{\eta}(r) - f_{\eta}(d))\| - \gamma)\\
\end{split}
\end{align}
\par
The predictive performance is measured by Hits@n metrics for
$n = 10, 100$, macro and micro mean rank and area under ROC curve (AUC
ROC). For rank-based metrics, we calculate the score of
$C \sqsubseteq \exists R.D$ for every class $C$ from the test set and
for every $D$ from the set $\mathbf{C}$ of all classes (or subclasses
of a certain type, such as proteins or functions) and determine the
rank of a test axiom $C \sqsubseteq \exists R.D$.  For macro mean rank
and AUC ROC we consider all axioms from the test set whereas for micro
metrics we compute corresponding class-specific metrics averaging them
over all classes in the signature:
\begin{equation}
    micro\_MR = Mean(MR_C(\{C \sqsubseteq \exists R.D, \thinspace D \in \mathbf{C}\}))
\end{equation}

\begin{equation}
    micro\_AUC\_ROC = Mean(AUC\_ROC_C(\{C \sqsubseteq \exists R.D, \thinspace D \in \mathbf{C}\}))
\end{equation}
Additionally, we remove axioms represented in the train set and obtain
corresponding filtered metrics (FHits@n, FMR, FAUC).

\section{Results}

Geometric methods such as {\it ELEmbeddings} address the task of knowledge
base completion by constructing a single (approximate) model for a
knowledge base and determining the truth of statements in this model
based on geometric scoring functions. However, a single model does not
suffice to compute entailments, or approximate
entailments. In first order logic or more expressive Description
Logics, it is possible to reduce entailment to the task of finding a
single model, but since \el does not allow for explicit negation, this
approach does not work; furthermore, reducing entailment to
consistency (i.e., not having a model) relies on solving an
optimization problem (``finding a model'') to compute each
entailment. Therefore, geometric methods only construct a single
model; the assumption is that any entailed statement has to be true in
this model, and some non-entailed statements will also be true. The
success of this approach relies on the model being sufficiently
expressive, and not constructing ``trivial'' models of knowledge
bases.

\begin{table}
\caption{{\it ELEmbeddings} experiments: the first column corresponds to the original model, the second one -- to LeakyReLU replacement and soft regularization, the third -- to GCI0-GCI3 losses added, and, finally, the last one -- to negatives filtering. {\it iw} refers to $interacts\_with$ dataset, {\it hf} -- to $has\_function$ dataset.}\label{tab:results}
\centering
\begin{tabular}{| C{1.5cm} | C{2cm} | C{1.7cm} | C{1.7cm} | C{1.7cm} | C{1.7cm} |}
\hline
& & ReLU & Leaky+Reg & Losses & Neg. filter \\
\hline
Yeast iw & FHits@10 & 0.00 & 0.26 & 0.25 & {\bf 0.29} \\
& FHits@100 & 0.15 & 0.74 & 0.74 & {\bf 0.78} \\
& macro\_FMR & 287.06 & 182.93 & 185.81 & {\bf 172.80} \\
& macro\_FAUC & 0.95 & {\bf 0.97} & {\bf 0.97} & {\bf 0.97} \\
\hline
\hline
Yeast hf & FHits@10 & 0.00 & {\bf 0.25} & 0.23 & 0.24 \\
& FHits@100 & 0.00 & {\bf 0.55} & 0.54 & 0.54 \\
& macro\_FMR & 5183.01 & 3211.80 & {\bf 2869.43} & 2875.67 \\
& macro\_FAUC & 0.90 & {\bf 0.94} & {\bf 0.94} & {\bf 0.94} \\
\hline
\hline
Human iw & FHits@10 & 0.00 & {\bf 0.02} & 0.00 & 0.00 \\
& FHits@100 & 0.03 & 0.24 & 0.50 & {\bf 0.63} \\
& macro\_FMR & 490.09 & 1361.12 & 258.17 & {\bf 196.76} \\
& macro\_FAUC & 0.97 & 0.93 & {\bf 0.99} & {\bf 0.99} \\
\hline
\hline
Human hf & FHits@10 & 0.00 & {\bf 0.14} & 0.06 & 0.06 \\
& FHits@100 & 0.00 & {\bf 0.35} & 0.28 & 0.28 \\
& macro\_FMR & 7642.15 & 4059.81 & 2270.35 & {\bf 2261.06} \\
& macro\_FAUC & 0.85 & 0.92 & {\bf 0.95} & {\bf 0.95} \\
\hline
\end{tabular}
\end{table}

\begin{table}
\caption{Naive approach vs {\it ELEmbeddings} with LeakyReLU, soft regularization constraints, GCI0-GCI3 negative losses and filtered negatives. {\it iw} refers to $interacts\_with$ dataset, {\it hf} -- to $has\_function$ dataset. For Human iw dataset we report here metrics for $M'_{iw}$, {\it sym} here corresponds to the symmetric Human iw dataset (for further details see Appendix (section~\ref{app:naive_model}).}\label{naive}
\centering
\begin{tabular}{| C{2.5cm} | C{2cm} | C{1.3cm} | C{1.3cm} |}
\hline
& & Naive & ELEm \\
\hline
Yeast iw & FHits@10 & 0.05 & {\bf 0.29} \\
& FHits@100 & 0.23 & {\bf 0.78} \\
& macro\_FMR & 1174.33 & {\bf 172.80} \\
& macro\_FAUC & 0.81 & {\bf 0.97} \\
\hline
\hline
Yeast hf & FHits@10 & 0.21 & {\bf 0.24} \\
& FHits@100 & 0.41 & {\bf 0.54} \\
& macro\_FMR & {\bf 2690.58} & 2875.67 \\
& macro\_FAUC & {\bf 0.95} & 0.94 \\
\hline
\hline
Human iw (sym) & FHits@10 & {\bf 0.02} & 0.00 \\
& FHits@100 & 0.08 & {\bf 0.63} \\
& macro\_FMR & 2299.09 & {\bf 196.76} \\
& macro\_FAUC & 0.88 & {\bf 0.99} \\
\hline
\hline
Human hf & FHits@10 & {\bf 0.18} & 0.06 \\
& FHits@100 & {\bf 0.39} & 0.28 \\
& macro\_FMR & {\bf 1967.45} & 2261.06 \\
& macro\_FAUC & {\bf 0.96} & 0.95 \\
\hline
\end{tabular}
\end{table}

\begin{figure}
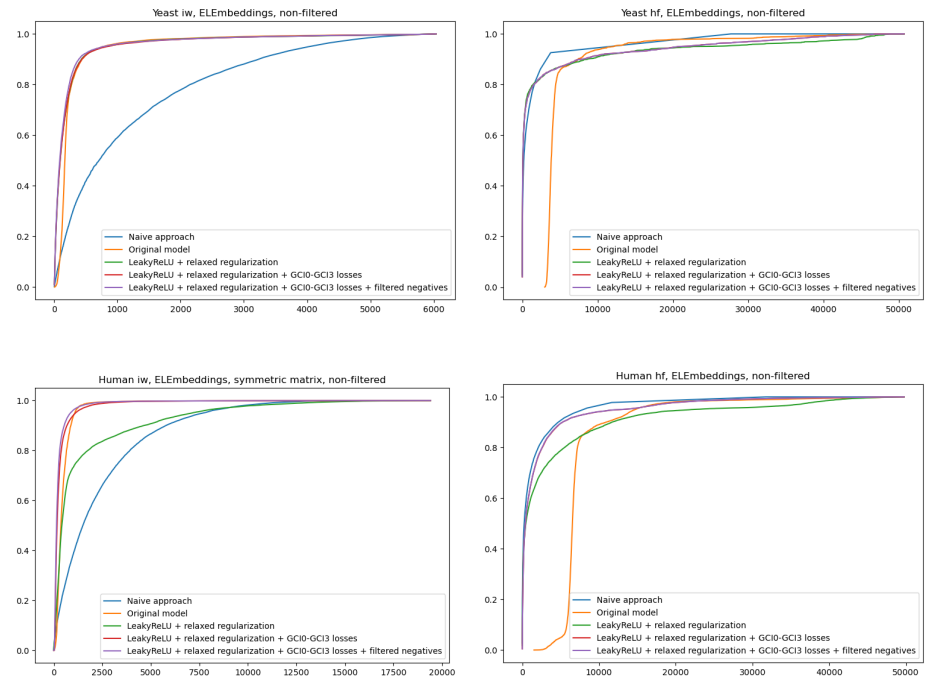

\begin{subfigure}{0.49\textwidth}
     \includegraphics[width=\textwidth]{figures/yeast_iw_roc_naive.png}
     \label{fig:aa}
 \end{subfigure}
 \hfill
 \begin{subfigure}{0.49\textwidth}
     \includegraphics[width=\textwidth]{figures/yeast_hf_roc_naive.png}
     \label{fig:bb}
 \end{subfigure}
 
 \medskip
 \begin{subfigure}{0.49\textwidth}
     \includegraphics[width=\textwidth]{figures/human_iw_roc_naive_sym.png}
     \label{fig:cc}
 \end{subfigure}
 \hfill
 \begin{subfigure}{0.49\textwidth}
     \includegraphics[width=\textwidth]{figures/human_hf_roc_naive.png}
     \label{fig:dd}
 \end{subfigure}
\caption{ROC curves across different models}
\label{fig:roc_curves}
\end{figure}

We use the {\it ELEmbeddings} method to perform knowledge base
completion in two applications which are used widely to benchmark
geometric ontology embedding methods, predicting protein--protein
interactions and predicting protein functions (see
Table~\ref{tab:results}; Appendix Figure~\ref{fig:roc_curves} shows
the resulting ROC curves).  To evaluate learned embeddings under
different modifications we run the original {\it ELEmbeddings} model
and use the obtained results instead of extracting metrics from the
original paper~\cite{kulmanov2019embeddings}. This is also motivated
by the utilization of different versions of GO and STRING database in
our work compared to the original paper~\cite{kulmanov2019embeddings}. We additionally perform an ablation study to evaluate the effect of individual modifications (see Appendix~\ref{app:ablation_study}).
We observe that {\it ELEmbeddings} ranks thousands of axioms at the
same rank (i.e., scores them as ``true''), and mainly achieves its
performance (measured in AUC) by ranking rare protein functions, or
proteins that interact rarely, at lower ranks. To further substantiate
this hypothesis, we developed a ``naive'' classifier that predicts
solely based on the number of times a class appears as part of an
axiom during training; Table~\ref{naive} shows the results and
demonstrates that only based on frequency of a class, a predictive
performance close to the actual performance of {\it ELEmbeddings} can
be achieved.

We first investigate whether a relaxation of the loss functions to
ensure non-zero gradients at all times can improve performance in the
knowledge base completion task. The loss functions are designed to
construct a model, and once an axiom from the knowledge base is true
in the constructed model, their losses remain zero; however, it may be
useful to provide a small gradient even once axioms are true in the
constructed model. For this purpose, we change the ReLU function used
in constructing losses to a LeakyReLU function. First, we study the
effect of replacing ReLU function with LeakyReLU together with
relaxed version of regularization (see Table~\ref{tab:results};
Appendix~\ref{app:full_el_embeddings_results} for full results). Since
LeakyReLU prevents gradients from being stuck at zero, we expect the
improvement of model's performance. Likewise, not forcing the centers
of $n$-balls representing concepts increases the expressiveness of the
model.  We demonstrate that, in general, incorporating LeakyReLU and
relaxing regularization improves the performance of the initial model
allowing learnable concepts to receive gradients at all times and, as
a consequence, construct a better approximate model. Furthermore, a
LeakyReLU adds the potential for optimization beyond ``truth'' (i.e.,
where statements are true in the constructed model and receive no
further updates that improve the task of knowledge base completion).

\begin{figure}
\begin{subfigure}{0.49\textwidth}
     \includegraphics[width=\textwidth]{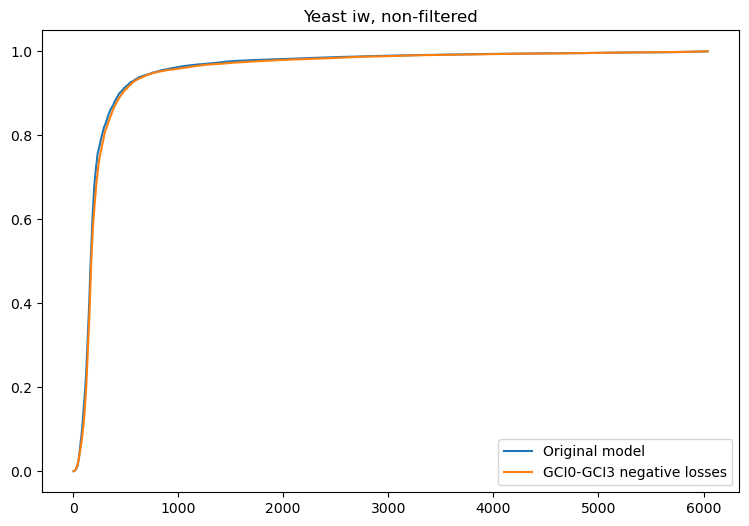}
     \label{fig:aaa}
 \end{subfigure}
 \hfill
 \begin{subfigure}{0.49\textwidth}
     \includegraphics[width=\textwidth]{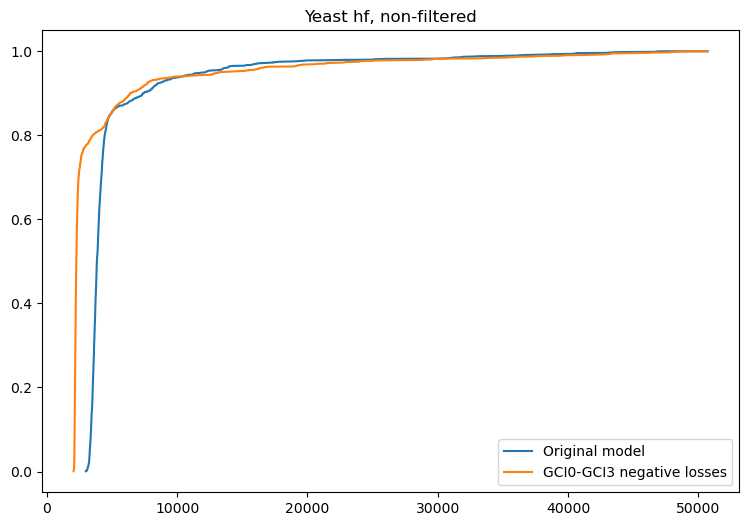}
     \label{fig:bbb}
 \end{subfigure}
 
 \medskip
 \begin{subfigure}{0.49\textwidth}
     \includegraphics[width=\textwidth]{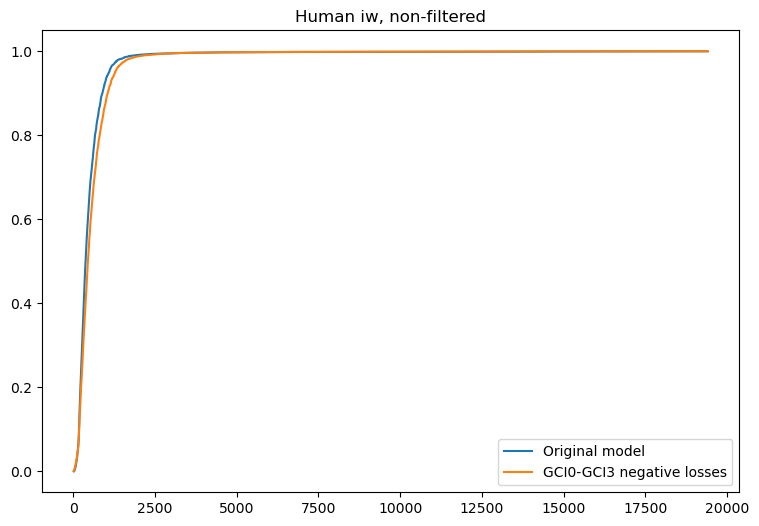}
     \label{fig:ccc}
 \end{subfigure}
 \hfill
 \begin{subfigure}{0.49\textwidth}
     \includegraphics[width=\textwidth]{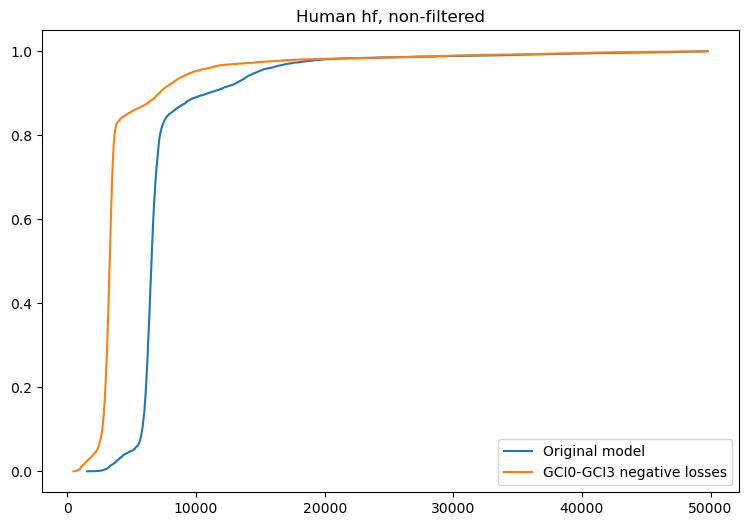}
     \label{fig:ddd}
 \end{subfigure}
\caption{{\it ELEmbeddings}, GCI0-GCI3 negative losses vs GCI2 negative loss}
\label{fig:losses_only}
\end{figure}

While the LeakyReLU improves the predictive performance of
{\it ELEmbeddings} in the task of knowledge base completion, it does not
prevent models from collapsing, i.e., generating trivial models (see
Appendix section~\ref{app:roc_curves_leaky_relu}). The original {\it
  ELEmbeddings} model and other geometric models only use negative
losses (i.e., losses for the case that an axiom does not hold) for a
single normal form (GCI2, $C \sqsubseteq \exists R.D$, which is also
used for prediction). We evaluate whether adding negative losses for
other normal forms will prevent the model from collapsing and improve
the performance in the task of knowledge base completion. We formulate
and add GCI0-GCI3 negative losses given by
equations~\ref{eq1}--\ref{eq3}, either separately or with LeakyReLU
and soft regularization from the previous experiment.  We find that
just adding the additional losses improves the performance and seems
to prevent models from collapsing (Appendix Figure~\ref{fig:losses_only}).
In terms of mean rank and AUC ROC, the
model with the negative losses generally exhibits improved performance
relative to using only negative losses for GCI2.

Similarly to how negative sampling works in knowledge graph
completion, geometric ontology embedding methods select negatives by
corrupting an axiom by replacing one of the classes with a randomly
chosen one; in the case of knowledge base completion where the
deductive closure contains potentially many non-trivial entailed
axioms, this approach may lead to suboptimal learning since some of
axioms treated as negatives are entailed (and will therefore be true
in any model, in particular the one constructed by the geometric
embedding method). We suggest to filter selected negatives based on
the deductive closure of the knowledge base: for each randomly
generated axiom to be used as negative, we check whether it is present
in the deductive closure and if it is, we delete it. To compute the
deductive closure, we use an approximate algorithm (see
Appendix~\ref{sec:closure}).  Table~\ref{tab:results} shows results in
the tasks we evaluate.  We find that excluding axioms in the deductive
closure for negative selection improves the results in the task of
predicting PPIs, and yields similar results in function prediction
tasks.  One possible reason is that a randomly chosen axiom is very
unlikely to be entailed since very few axioms are entailed compared to
all possible axioms to choose from.

\begin{figure}
\begin{subfigure}{0.49\textwidth}
     \includegraphics[width=\textwidth]{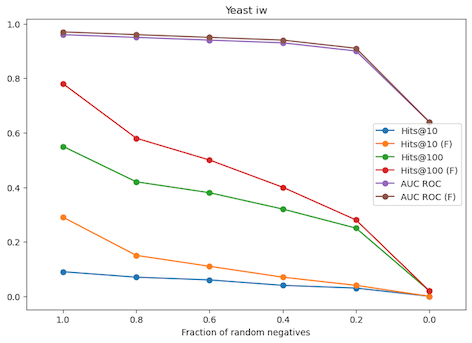}
     \label{fig:a}
 \end{subfigure}
 \hfill
 \begin{subfigure}{0.49\textwidth}
     \includegraphics[width=\textwidth]{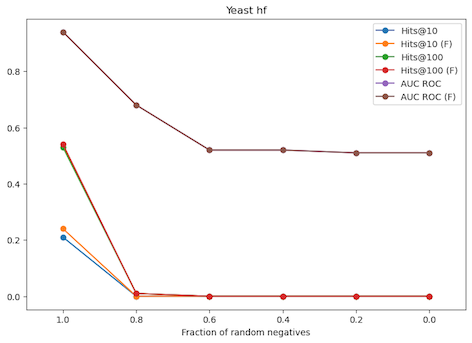}
     \label{fig:b}
 \end{subfigure}
 
 \medskip
 \begin{subfigure}{0.49\textwidth}
     \includegraphics[width=\textwidth]{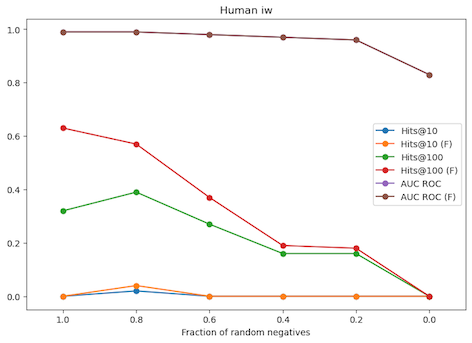}
     \label{fig:c}
 \end{subfigure}
 \hfill
 \begin{subfigure}{0.49\textwidth}
     \includegraphics[width=\textwidth]{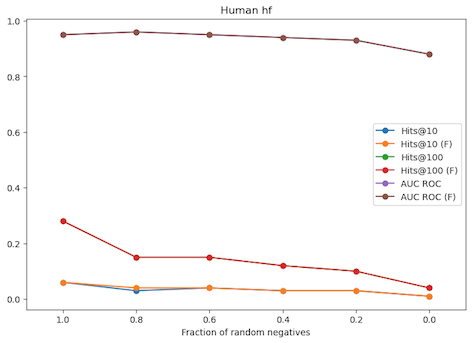}
     \label{fig:d}
 \end{subfigure}
\caption{Random negatives sampling}
\label{random_negatives}
\end{figure}

Because the chance of selecting an entailed axiom as a negative
depends on the knowledge base on which the embedding method is
applied, we perform additional experiments where we bias the selection
of negatives; we chose between 100\% negatives to 0\% negatives from
the entailed axioms.
We find that reducing the number of entailed axioms from the negatives
has an effect to improve performance and the effect increases the more
axioms would be chosen from the entailed ones (Appendix
Figure~\ref{random_negatives}).

The deductive closure can also be used to modify the evaluation
metrics. So far, ontology embedding methods that have been applied to
the task of knowledge base completion have used evaluation measures
that are taken from the task of knowledge graph completion; in
particular, they only evaluate knowledge base completion using axioms
that are ``novel'' and not entailed. However, any entailed axiom will
be true in all models of the knowledge base, and therefore also in the
geometric model that is constructed by the embedding method. These
entailed axioms should therefore be considered in the evaluation. We
show the difference in performance, and the corresponding ROC curves,
in Appendix Figure~\ref{fig:elem_dc}.
We find that methods that explicitly construct models generally
predict entailed axioms first, even when the models make some trivial
predictions (such as in the original {\it ELEmbeddings} model);
model-generating embedding first predict the entailed axioms, and then
predict ``novel'' axioms that are not entailed. However, when
replacing the ReLU with the LeakyReLU in {\it ELEmbeddings}, ``novel'',
non-entailed axioms are predicted first, before entailed axioms are
predicted (see Appendix Figure~\ref{fig:elem_leaky_dc}). We evaluate a
more recent ontology embedding method {\it
  $Box^2EL$}~\cite{jackermeier2023box} and find that this model
predicts primarily ``novel'' axioms but does not predict entailed
axioms (see Appendix Figure~\ref{fig:box2el_dc}).

\section{Discussion}

We evaluated properties of {\it ELEmbeddings}, an ontology embedding method
that aims to generate a model of an \el theory; the properties we
evaluate hold similarly for other ontology embedding methods that
construct models of \el theories. While we demonstrate several
improvements over the original model, we can also draw some general
conclusions about ontology embedding methods and their
evaluation. Knowledge base completion is the task of predicting axioms
that should be added to a knowledge base; this task is adapted from
knowledge graph completion where triples are added to a knowledge
graph. The way both tasks are evaluated is by removing some statements
(axioms or triples) from the knowledge base, and evaluating whether
these axioms or triples can be recovered by the embedding method. This
evaluation approach is adequate for knowledge graphs which do not give
rise to many entailments. However, knowledge bases give rise to
potentially many non-trivial entailments that need to be considered in
the evaluation. In particular embedding methods that aim to generate a
model of a knowledge base will first generate entailed axioms (because
entailed axioms are true in all models); these methods perform
knowledge base completion as a generalization of generating the model
where either other statements may be true, or they may be
approximately true in the generated structure. This has two
consequences: the evaluation procedure needs to account for this; and
the model needs to be sufficiently rich to allow useful predictions.

We have introduced a method to compute the deductive closure of \el
know\-ledge bases; this method relies on an automated reasoner and is
sound but not complete. We use all the axioms in the deductive closure
as positive axioms to be predicted when evaluating knowledge base
completion, to account for methods that treat knowledge base
completion as a generalization of constructing a model and testing for
truth in this model. We find that some models (e.g., the original
ELEmbedding model) can predict entailed axioms well, some (a
modified model using a LeakyReLU function as part of the loss instead
of the ReLU) preferentially predict ``novel'', non-entailed axioms,
and others (e.g., the {\it $Box2EL$} model) are tailored to predict
primarily ``novel'' knowledge and do not predict entailed axioms;
these methods solve subtly different problems (either generalizing
construction of a model, or specifically predicting novel non-entailed
axioms). 
We also modify the evaluation procedure to account for the
inclusion of entailed axioms as positives; however, the evaluation
measures are still based on ranking individual axioms and do not
account for semantic similarity. For example, if during testing, the
correct axiom to predict is $C \sqsubseteq \exists R.D$ but the
predicted axiom is $C \sqsubseteq \exists R.E$, the prediction may be
considered to be ``more correct'' if $D \sqsubseteq E$ was in the
knowledge base than if $D \sqcap E \sqsubseteq \bot$ was in the
knowledge base. Novel evaluation metrics need to be designed to
account for this phenomenon, similarly to ontology-based evaluation
measures used in life sciences \cite{radivojac2013}. It is also
important to expand the set of benchmark sets for knowledge base
completion.

Use of the deductive closure is not only useful in evaluation but also
when selecting negatives. In formal knowledge bases, there are at
least two ways in which negatives for axioms can be chosen: they are
either non-entailed axioms, or they are axioms whose negation is
entailed. However, in no case should entailed axioms be considered as
negatives; we demonstrate that filtering entailed axioms from selected
negatives during training improves the performance of the embedding
method consistently in knowledge base completion (and, obviously, more
so when entailed axioms are considered as positives during
evaluation).

While we only report our experiments with {\it ELEmbeddings}, our findings,
in particular about the evaluation and use of deductive closure, 
are applicable to other geometric ontology embedding methods. As
ontology embedding methods are increasingly applied in
knowledge-enhanced learning and other tasks that utilize some form of
approximate computation of entailments, our results can also serve to
improve the applications of ontology embeddings.

\bibliographystyle{splncs04}
\bibliography{refs}

\newpage
\appendix

\section*{Appendix}

\section{Naive model construction}~\label{app:naive_model}

Similarly to~\cite{hinnerichs2021dti} we construct $n \times n$ PPI
and $n \times m$ function prediction matrices $M_{iw}$ and $M_{hf}$
respectively: $M_{iw}(P_1, P_2) = 1$ if
$\{P_1\} \sqsubseteq \exists interacts\_with.\{P_2\}$ is in the train set for PPI and $0$ otherwise, and $M_{hf}(P, GO) = 1$ if $\{P\} \sqsubseteq \exists has\_function.\{GO\}$ is in the train set for function prediction ($0$ otherwise). Assuming that $interacts\_with$ relation is symmetric we additionally design matrix $M'_{iw}$ for human data where $M'_{iw}(P_1, P_2) = M'_{iw}(P_2, P_1) = 1$ when $\{P_1\} \sqsubseteq \exists interacts\_with.\{P_2\}$ or $\{P_2\} \sqsubseteq \exists interacts\_with.\{P_1\}$ can be found in the train part of the dataset. Scoring function used for rank-based predictions is described in section~\ref{training_procedure}.

\section{GCI statistics}~\label{app:gci_numbers}

\begin{table}[h!]
\caption{Datasets' statistics}\label{tab:dataset-stat}
\centering
\begin{tabular}{|l|l|r|}
\hline
Dataset name & GCI type & Number of axioms \\
\hline
Yeast & $C \sqsubseteq D$ & 81,068 \\
      & $C \sqsubseteq \bot$ & 0 \\
      & $C \sqcap D \sqsubseteq E$ & 11,825 \\
      & $C \sqcap D \sqsubseteq \bot$ & 31 \\
      & $C \sqsubseteq \exists R.D$ & 293,645 \\
      & $\exists R.C \sqsubseteq D$ & 11,823 \\
      & $\exists R.C \sqsubseteq \bot$ & 0 \\
\hline
Human & $C \sqsubseteq D$ & 87,555 \\
      & $C \sqsubseteq \bot$ & 0 \\
      & $C \sqcap D \sqsubseteq E$ & 12,154 \\
      & $C \sqcap D \sqsubseteq \bot$ & 30 \\
      & $C \sqsubseteq \exists R.D$ & 958,674 \\
      & $\exists R.C \sqsubseteq D$ & 12,152 \\
      & $\exists R.C \sqsubseteq \bot$ & 0 \\
\hline
\end{tabular}
\end{table}

\newpage 

\section{Hyperparameters}~\label{sec:hyper}

\begin{table}[h!]
\caption{Best models' hyperparameters, Yeast iw dataset}\label{tab:yeast_iw_hyperparameters}
\centering
\begin{tabular}{|c|c|}
\hline
Experiment & Best hyperparameters \\
\hline
{\it ELEmbeddings} original & embed\_dim = 200 \\
                            & $\gamma = 0.1$ \\
                            & learning\_rate = 0.01 \\
\hline
LeakyReLU + relaxed regularization & embed\_dim = 50 \\
                                   & $\gamma = 0.1$ \\
                                   & $R = 1$ \\
                                   & learning\_rate = 0.01 \\
\hline
GCI0-GCI3 negative losses & embed\_dim = 50 \\
                          & $\gamma = 0.1$ \\
                          & $R = 2$ \\
                          & learning\_rate = 0.01 \\
\hline
Filtered negatives & embed\_dim = 100 \\
                   & $\gamma = 0.1$ \\
                   & $R = 1$ \\
                   & learning\_rate = 0.01 \\
\hline
\end{tabular}
\end{table}

\begin{table}[h!]
\caption{Best models' hyperparameters, Yeast hf dataset}\label{tab:yeast_hf_hyperparameters}
\centering
\begin{tabular}{|c|c|}
\hline
Experiment & Best hyperparameters \\
\hline
{\it ELEmbeddings} original & embed\_dim = 400 \\
                            & $\gamma = 0.1$ \\
                            & learning\_rate = 0.001 \\
\hline
LeakyReLU + relaxed regularization & embed\_dim = 100 \\
                                   & $\gamma = 0.1$ \\
                                   & $R = 2$ \\
                                   & learning\_rate = 0.0001 \\
\hline
GCI0-GCI3 negative losses & embed\_dim = 200 \\
                          & $\gamma = 0.1$ \\
                          & $R = 1$ \\
                          & learning\_rate = 0.0001 \\
\hline
Filtered negatives & embed\_dim = 200 \\
                   & $\gamma = 0.1$ \\
                   & $R = 1$ \\
                   & learning\_rate = 0.0001 \\
\hline
\end{tabular}
\end{table}

\begin{table}[h!]
\caption{Best models' hyperparameters, Human iw dataset}\label{tab:human_iw_hyperparameters}
\centering
\begin{tabular}{|c|c|}
\hline
Experiment & Best hyperparameters \\
\hline
{\it ELEmbeddings} original & embed\_dim = 400 \\
                            & $\gamma = 0.1$ \\
                            & learning\_rate = 0.001 \\
\hline
LeakyReLU + relaxed regularization & embed\_dim = 100 \\
                                   & $\gamma = 0.01$ \\
                                   & $R = 1$ \\
                                   & learning\_rate = 0.0001 \\
\hline
GCI0-GCI3 negative losses & embed\_dim = 200 \\
                          & $\gamma = 0.1$ \\
                          & $R = 2$ \\
                          & learning\_rate = 0.001 \\
\hline
Filtered negatives & embed\_dim = 200 \\
                   & $\gamma = 0.1$ \\
                   & $R = 2$ \\
                   & learning\_rate = 0.001 \\
\hline
\end{tabular}
\end{table}

\begin{table}[h!]
\caption{Best models' hyperparameters, Human hf dataset}\label{tab:human_hf_hyperparameters}
\centering
\begin{tabular}{|c|c|}
\hline
Experiment & Best hyperparameters \\
\hline
{\it ELEmbeddings} original & embed\_dim = 50 \\
                            & $\gamma = 0.1$ \\
                            & learning\_rate = 0.001 \\
\hline
LeakyReLU + relaxed regularization & embed\_dim = 50 \\
                                   & $\gamma = 0.1$ \\
                                   & $R = 2$ \\
                                   & learning\_rate = 0.0001 \\
\hline
GCI0-GCI3 negative losses & embed\_dim = 400 \\
                          & $\gamma = 0.1$ \\
                          & $R = 2$ \\
                          & learning\_rate = 0.0001 \\
\hline
Filtered negatives & embed\_dim = 400 \\
                   & $\gamma = 0.1$ \\
                   & $R = 2$ \\
                   & learning\_rate = 0.0001 \\
\hline
\end{tabular}
\end{table}

\section{Ablation study hyperparameters}~\label{app:ablation-hyper}

\begin{table}[h!]
\caption{Best models' hyperparameters for ablation study, Yeast iw dataset}\label{tab:yeast_iw_ablation_hyperparameters}
\centering
\begin{tabular}{|c|c|}
\hline
Experiment & Best hyperparameters \\
\hline
LeakyReLU & embed\_dim = 200 \\
          & $\gamma = 0.1$ \\
          & learning\_rate = 0.01 \\
\hline
GCI0-GCI3 negative losses & embed\_dim = 100 \\
                          & $\gamma = 0.1$ \\
                          & learning\_rate = 0.01 \\
\hline
Relaxed regularization & embed\_dim = 100 \\
                       & $\gamma = 0.01$ \\
                       & $R = 1$ \\
                       & learning\_rate = 0.01 \\
\hline
Filtered negatives & embed\_dim = 400 \\
                   & $\gamma = 0.01$ \\
                   & learning\_rate = 0.01 \\
\hline
\end{tabular}
\end{table}

\begin{table}[h!]
\caption{Best models' hyperparameters for ablation study, Yeast hf dataset}\label{tab:yeast_hf_ablation_hyperparameters}
\centering
\begin{tabular}{|c|c|}
\hline
Experiment & Best hyperparameters \\
\hline
LeakyReLU & embed\_dim = 400 \\
          & $\gamma = -0.1$ \\
          & learning\_rate = 0.0001 \\
\hline
GCI0-GCI3 negative losses & embed\_dim = 200 \\
                          & $\gamma = 0.1$ \\
                          & learning\_rate = 0.01 \\
\hline
Relaxed regularization & embed\_dim = 200 \\
                       & $\gamma = 0.1$ \\
                       & $R = 1$ \\
                       & learning\_rate = 0.01 \\
\hline
Filtered negatives & embed\_dim = 400 \\
                   & $\gamma = 0.1$ \\
                   & learning\_rate = 0.001 \\
\hline
\end{tabular}
\end{table}

\begin{table}[h!]
\caption{Best models' hyperparameters for ablation study, Human iw dataset}\label{tab:human_iw_ablation_hyperparameters}
\centering
\begin{tabular}{|c|c|}
\hline
Experiment & Best hyperparameters \\
\hline
LeakyReLU & embed\_dim = 50 \\
          & $\gamma = -0.1$ \\
          & learning\_rate = 0.0001 \\
\hline
GCI0-GCI3 negative losses & embed\_dim = 200 \\
                          & $\gamma = 0.1$ \\
                          & learning\_rate = 0.001 \\
\hline
Relaxed regularization & embed\_dim = 400 \\
                       & $\gamma = 0.1$ \\
                       & $R = 1$ \\
                       & learning\_rate = 0.001 \\
\hline
Filtered negatives & embed\_dim = 400 \\
                   & $\gamma = 0.1$ \\
                   & learning\_rate = 0.001 \\
\hline
\end{tabular}
\end{table}

\begin{table}[h!]
\caption{Best models' hyperparameters for ablation study, Human hf dataset}\label{tab:human_hf_ablation_hyperparameters}
\centering
\begin{tabular}{|c|c|}
\hline
Experiment & Best hyperparameters \\
\hline
LeakyReLU & embed\_dim = 50 \\
          & $\gamma = 0.01$ \\
          & learning\_rate = 0.01 \\
\hline
GCI0-GCI3 negative losses & embed\_dim = 400 \\
                          & $\gamma = 0.1$ \\
                          & learning\_rate = 0.001 \\
\hline
Relaxed regularization & embed\_dim = 400 \\
                       & $\gamma = 0.01$ \\
                       & $R = 2$ \\
                       & learning\_rate = 0.01 \\
\hline
Filtered negatives & embed\_dim = 50 \\
                   & $\gamma = 0.1$ \\
                   & learning\_rate = 0.001 \\
\hline
\end{tabular}
\end{table}

\section{Detailed results of different settings of {\it ELEmbeddings}}~\label{app:full_el_embeddings_results}
{\it ELEmbeddings} experiments: the first column corresponds to the original model, the second one -- to LeakyReLU replacement and soft regularization, the third -- to GCI0-GCI3 losses added, and, finally, the last one -- to negatives filtering. {\it iw} refers to $interacts\_with$ dataset, {\it hf} -- to $has\_function$ dataset.

\begin{table}[h!]
\centering
\begin{tabular}{| C{1.5cm} | C{2cm} | C{1.7cm} | C{1.7cm} | C{1.7cm} | C{1.7cm} |}
\hline
& & ReLU & Leaky+Reg & Losses & Neg. filter \\
\hline
Yeast iw & Hits@10 & 0.00 & {\bf 0.09} & {\bf 0.09} & {\bf 0.09} \\
         & FHits@10 & 0.00 & 0.26 & 0.25 & {\bf 0.29} \\
         & Hits@100 & 0.15 & 0.52 & 0.52 & {\bf 0.55} \\
         & FHits@100 & 0.15 & 0.74 & 0.74 & {\bf 0.78} \\
         & macro\_MR & 287.36 & 242.76 & 245.21 & {\bf 231.70} \\
         & micro\_MR & 359.70 & 310.93 & 310.46 & {\bf 296.12} \\
         & macro\_FMR & 287.06 & 182.93 & 185.81 & {\bf 172.80} \\
         & micro\_FMR & 359.61 & 280.76 & 280.44 & {\bf 266.51} \\
         & macro\_AUC & 0.95 & {\bf 0.96} & {\bf 0.96} & {\bf 0.96} \\
         & micro\_AUC & 0.95 & {\bf 0.96} & {\bf 0.96} & {\bf 0.96} \\
         & macro\_FAUC & 0.95 & {\bf 0.97} & {\bf 0.97} & {\bf 0.97} \\
         & micro\_FAUC & 0.95 & {\bf 0.97} & 0.96 & 0.96 \\
\hline
\end{tabular}
\end{table}

\begin{table}[h!]
\centering
\begin{tabular}{| C{1.5cm} | C{2cm} | C{1.7cm} | C{1.7cm} | C{1.7cm} | C{1.7cm} |}
\hline
& & ReLU & Leaky+Reg & Losses & Neg. filter \\
\hline
Yeast hf & Hits@10 & 0.00 & {\bf 0.22} & 0.21 & 0.21 \\
         & FHits@10 & 0.00 & {\bf 0.25} & 0.23 & 0.24 \\
         & Hits@100 & 0.00 & {\bf 0.54} & {\bf 0.54} & 0.53 \\
         & FHits@100 & 0.00 & {\bf 0.55} & 0.54 & 0.54 \\
         & macro\_MR & 5183.02 & 3215.76 & {\bf 2873.35} & 2879.59 \\
         & micro\_MR & 5205.98 & 3174.81 & {\bf 2850.97} & 2858.37 \\
         & macro\_FMR & 5183.01 & 3211.80 & {\bf 2869.43} & 2875.67 \\
         & micro\_FMR & 5205.97 & 3171.08 & {\bf 2847.28} & 2854.69 \\
         & macro\_AUC & 0.90 & {\bf 0.94} & {\bf 0.94} & {\bf 0.94} \\
         & micro\_AUC & 0.90 & 0.94 & {\bf 0.95} & {\bf 0.95} \\
         & macro\_FAUC & 0.90 & {\bf 0.94} & {\bf 0.94} & {\bf 0.94} \\
         & micro\_FAUC & 0.90 & 0.94 & {\bf 0.95} & {\bf 0.95} \\
\hline
\end{tabular}
\end{table}

\begin{table}[h!]
\centering
\begin{tabular}{| C{1.5cm} | C{2cm} | C{1.7cm} | C{1.7cm} | C{1.7cm} | C{1.7cm} |}
\hline
& & ReLU & Leaky+Reg & Losses & Neg. filter \\
\hline
Human iw & Hits@10 & 0.00 & {\bf 0.01} & 0.00 & 0.00 \\
         & FHits@10 & 0.00 & {\bf 0.02} & 0.00 & 0.00 \\
         & Hits@100 & 0.03 & 0.12 & 0.26 & {\bf 0.32} \\
         & FHits@100 & 0.03 & 0.24 & 0.50 & {\bf 0.63} \\
         & macro\_MR & 490.14 & 1440.09 & 338.67 & {\bf 277.22} \\
         & micro\_MR & 513.98 & 1973.29 & 339.97 & {\bf 283.61} \\
         & macro\_FMR & 490.09 & 1361.12 & 258.17 & {\bf 196.76} \\
         & micro\_FMR & 513.97 & 1936.41 & 302.18 & {\bf 245.83} \\
         & macro\_AUC & 0.97 & 0.93 & 0.98 & 0.99 \\
         & micro\_AUC & 0.97 & 0.91 & 0.98 & 0.99 \\
         & macro\_FAUC & 0.97 & 0.93 & {\bf 0.99} & {\bf 0.99} \\
         & micro\_FAUC & 0.97 & 0.91 & {\bf 0.99} & {\bf 0.99} \\
\hline
\end{tabular}
\end{table}

\begin{table}[h!]
\centering
\begin{tabular}{| C{1.5cm} | C{2cm} | C{1.7cm} | C{1.7cm} | C{1.7cm} | C{1.7cm} |}
\hline
& & ReLU & Leaky+Reg & Losses & Neg. filter \\
\hline
Human hf & Hits@10 & 0.00 & {\bf 0.13} & 0.06 & 0.06 \\
         & FHits@10 & 0.00 & {\bf 0.14} & 0.06 & 0.06 \\
         & Hits@100 & 0.00 & {\bf 0.34} & 0.28 & 0.28 \\
         & FHits@100 & 0.00 & {\bf 0.35} & 0.28 & 0.28 \\
         & macro\_MR & 7642.19 & 4070.90 & 2281.62 & {\bf 2272.33} \\
         & micro\_MR & 7645.75 & 3578.47 & 1972.00 & {\bf 1963.87} \\
         & macro\_FMR & 7642.15 & 4059.81 & 2270.35 & {\bf 2261.06} \\
         & micro\_FMR & 7645.73 & 3570.33 & 1963.76 & {\bf 1955.62} \\
         & macro\_AUC & 0.85 & 0.92 & {\bf 0.95} & {\bf 0.95} \\
         & micro\_AUC & 0.85 & 0.94 & {\bf 0.97} & {\bf 0.97} \\
         & macro\_FAUC & 0.85 & 0.92 & {\bf 0.95} & {\bf 0.95} \\
         & micro\_FAUC & 0.85 & 0.94 & {\bf 0.97} & {\bf 0.97} \\
\hline
\end{tabular}
\end{table}

\section{Detailed results of comparison with the ``naive'' classifier.}~\label{app:results_with_naive}
Naive approach vs {\it ELEmbeddings} with LeakyReLU, soft regularization constraints, GCI0-GCI3 negative losses and filtered negatives. {\it iw} refers to $interacts\_with$ dataset, {\it hf} -- to $has\_function$ dataset. For Human iw dataset we report here metrics both for $M'_{iw}$ and $M_{iw}$, {\it sym} here corresponds to the symmetric Human iw dataset (for further details see Appendix (section~\ref{app:naive_model}).

\begin{table}[h!]
\centering
\begin{tabular}{| C{3.5cm} | C{2cm} | C{1.3cm} | C{1.3cm} |}
\hline
& & Naive & ELEm \\
\hline
Yeast iw & Hits@10 & 0.01 & {\bf 0.09} \\
         & FHits@10 & 0.05 & {\bf 0.29} \\
         & Hits@100 & 0.12 & {\bf 0.55} \\
         & FHits@100 & 0.23 & {\bf 0.78} \\
         & macro\_MR & 1228.68 & {\bf 231.70} \\
         & micro\_MR & 1845.96 & {\bf 296.12} \\
         & macro\_FMR & 1174.33 & {\bf 172.80} \\
         & micro\_FMR & 1819.47 & {\bf 266.51} \\
         & macro\_AUC & 0.80 & {\bf 0.96} \\
         & micro\_AUC & 0.72 & {\bf 0.96} \\
         & macro\_FAUC & 0.81 & {\bf 0.97} \\
         & micro\_FAUC & 0.72 & {\bf 0.96} \\
\hline
\end{tabular}
\end{table}

\begin{table}[h!]
\centering
\begin{tabular}{| C{3.5cm} | C{2cm} | C{1.3cm} | C{1.3cm} |}
\hline
& & Naive & ELEm \\
\hline
Yeast hf & Hits@10 & 0.20 & {\bf 0.21} \\
         & FHits@10 & 0.21 & {\bf 0.24} \\
         & Hits@100 & 0.40 & {\bf 0.53} \\
         & FHits@100 & 0.41 & {\bf 0.54} \\
         & macro\_MR & {\bf 2694.33} & 2879.59 \\
         & micro\_MR & {\bf 2658.51} & 2858.37 \\
         & macro\_FMR & {\bf 2690.58} & 2875.67 \\
         & micro\_FMR & {\bf 2654.98} & 2854.69 \\
         & macro\_AUC & {\bf 0.97} & 0.94 \\
         & micro\_AUC & 0.95 & 0.95 \\
         & macro\_FAUC & {\bf 0.95} & 0.94 \\
         & micro\_FAUC & 0.95 & 0.95 \\
\hline
\end{tabular}
\end{table}

\begin{table}[h!]
\centering
\begin{tabular}{| C{3.5cm} | C{2cm} | C{1.3cm} | C{1.3cm} |}
\hline
& & Naive & ELEm \\
\hline
Human iw (sym) & Hits@10 & {\bf 0.01} & 0.00 \\
               & FHits@10 & {\bf 0.02} & 0.00 \\
               & Hits@100 & 0.07 & {\bf 0.32} \\
               & FHits@100 & 0.08 & {\bf 0.63} \\
               & macro\_MR & 2377.39 & {\bf 277.22} \\
               & micro\_MR & 3313.50 & {\bf 283.61} \\
               & macro\_FMR & 2299.09 & {\bf 196.76} \\
               & micro\_FMR & 3276.97 & {\bf 245.83} \\
               & macro\_AUC & 0.88 & {\bf 0.99} \\
               & micro\_AUC & 0.84 & {\bf 0.99} \\
               & macro\_FAUC & 0.88 & {\bf 0.99} \\
               & micro\_FAUC & 0.85 & {\bf 0.99} \\
\hline
\end{tabular}
\end{table}

\begin{table}[h!]
\centering
\begin{tabular}{| C{3.5cm} | C{2cm} | C{1.3cm} | C{1.3cm} |}
\hline
& & Naive & ELEm \\
\hline
Human iw (non-sym) & Hits@10 & {\bf 0.01} & 0.00 \\
                   & FHits@10 & {\bf 0.02} & 0.00 \\
                   & Hits@100 & 0.07 & {\bf 0.32} \\
                   & FHits@100 & 0.08 & {\bf 0.63} \\
                   & macro\_MR & 2433.80 & {\bf 277.22} \\
                   & micro\_MR & 3412.09 & {\bf 283.61} \\
                   & macro\_FMR & 2353.87 & {\bf 196.76} \\
                   & micro\_FMR & 3374.88 & {\bf 245.83} \\
                   & macro\_AUC & 0.88 & {\bf 0.99} \\
                   & micro\_AUC & 0.84 & {\bf 0.99} \\
                   & macro\_FAUC & 0.88 & {\bf 0.99} \\
                   & micro\_FAUC & 0.84 & {\bf 0.99} \\
\hline
\end{tabular}
\end{table}

\begin{table}[h!]
\centering
\begin{tabular}{| C{3.5cm} | C{2cm} | C{1.3cm} | C{1.3cm} |}
\hline
& & Naive & ELEm \\
\hline
Human hf & Hits@10 & {\bf 0.16} & 0.06 \\
         & FHits@10 & {\bf 0.18} & 0.06 \\
         & Hits@100 & {\bf 0.39} & 0.28 \\
         & FHits@100 & {\bf 0.39} & 0.28 \\
         & macro\_MR & {\bf 1978.53} & 2272.33 \\
         & micro\_MR & {\bf 1785.81} & 1963.87 \\
         & macro\_FMR & {\bf 1967.45} & 2261.06 \\
         & micro\_FMR & {\bf 1777.73} & 1955.62 \\
         & macro\_AUC & {\bf 0.97} & {\bf 0.95} \\
         & micro\_AUC & 0.97 & {\bf 0.97} \\
         & macro\_FAUC & {\bf 0.96} & 0.95 \\
         & micro\_FAUC & 0.97 & {\bf 0.97} \\
\hline
\end{tabular}
\end{table}

Note that by definition filtered metrics should be less than or equal
to corresponding non-filtered metrics, yet here filtered naive AUC ROC
is less than non-filtered one. The reason is trapezoidal rule for
numerical integration used to calculate AUC ROC based on FPR and TPR
points: due to the facts that the number of different rank values is
relatively small compared to the number of GO classes and that the
score relies exclusively on the number of proteins having the
function, it provides the grid not accurate enough, and the same
non-filtered rank converts into multiple lower ranks while filtered
forming more well-suited computational grid.

\section{Ablation study}~\label{app:ablation_study}
{\it ELEmbeddings} experiments: the first column corresponds to the original model, the second one -- to LeakyReLU replacement, the third one -- to soft regularization, the fourth -- to GCI0-GCI3 losses, and, finally, the last one -- to negatives filtering. {\it iw} refers to $interacts\_with$ dataset, {\it hf} -- to $has\_function$ dataset. Best hyperparameters for these experiments can be found in Appendix~\ref{app:ablation-hyper}.

\begin{table}[h!]
\centering
\begin{tabular}{| C{1.5cm} | C{2cm} | C{1.7cm} | C{1.7cm} | C{1.7cm} | C{1.7cm} | C{1.7cm} |}
\hline
& & ReLU & LeakyReLU & Reg & Losses & Neg. filter \\
\hline
Yeast iw & Hits@10 & 0.00 & 0.09 & 0.00 & 0.00 & 0.00 \\
         & FHits@10 & 0.00 & 0.27 & 0.00 & 0.00 & 0.00 \\
         & Hits@100 & 0.15 & 0.54 & 0.11 & 0.14 & 0.11 \\
         & FHits@100 & 0.15 & 0.76 & 0.11 & 0.14 & 0.11 \\
         & macro\_MR & 287.36 & 236.67 & 309.77 & 329.22 & 327.87 \\
         & micro\_MR & 359.70 & 301.79 & 378.44 & 409.94 & 415.26 \\
         & macro\_FMR & 287.06 & 176.64 & 309.59 & 328.86 & 327.87 \\
         & micro\_FMR & 359.61 & 271.57 & 378.38 & 409.83 & 415.26 \\
         & macro\_AUC & 0.95 & 0.96 & 0.95 & 0.95 & 0.95 \\
         & micro\_AUC & 0.95 & 0.96 & 0.94 & 0.94 & 0.94 \\
         & macro\_FAUC & 0.95 & 0.97 & 0.95 & 0.95 & 0.95 \\
         & micro\_FAUC & 0.95 & 0.96 & 0.94 & 0.94 & 0.94 \\
\hline
\end{tabular}
\end{table}

\begin{table}[h!]
\centering
\begin{tabular}{| C{1.5cm} | C{2cm} | C{1.7cm} | C{1.7cm} | C{1.7cm} | C{1.7cm} | C{1.7cm} |}
\hline
& & ReLU & LeakyReLU & Reg & Losses & Neg. filter \\
\hline
Yeast hf & Hits@10 & 0.00 & 0.25 & 0.00 & 0.00 & 0.00 \\
         & FHits@10 & 0.00 & 0.28 & 0.00 & 0.00 & 0.00 \\
         & Hits@100 & 0.00 & 0.54 & 0.00 & 0.00 & 0.00 \\
         & FHits@100 & 0.00 & 0.55 & 0.00 & 0.00 & 0.00 \\
         & macro\_MR & 5183.02 & 2770.44 & 8252.35 & 3450.22 & 5287.68 \\
         & micro\_MR & 5205.98 & 2752.54 & 8221.89 & 3426.88 &  5311.96 \\
         & macro\_FMR & 5183.01 & 2766.28 & 8252.33 & 3450.21 & 5287.67 \\
         & micro\_FMR & 5205.97 & 2748.63 & 8221.88 & 3426.87 & 5311.95 \\
         & macro\_AUC & 0.90 & 0.95 & 0.84 & 0.93 & 0.90 \\
         & micro\_AUC & 0.90 & 0.95 & 0.84 & 0.93 & 0.90 \\
         & macro\_FAUC & 0.90 & 0.95 & 0.84 & 0.93 & 0.90 \\
         & micro\_FAUC & 0.90 & 0.95 & 0.84 & 0.93 & 0.90 \\
\hline
\end{tabular}
\end{table}

\begin{table}[h!]
\centering
\begin{tabular}{| C{1.5cm} | C{2cm} | C{1.7cm} | C{1.7cm} | C{1.7cm} | C{1.7cm} | C{1.7cm} |}
\hline
& & ReLU & LeakyReLU & Reg & Losses & Neg. filter \\
\hline
Human iw & Hits@10 & 0.00 & 0.03 & 0.00 & 0.00 & 0.00 \\
         & FHits@10 & 0.00 & 0.09 & 0.00 & 0.00 & 0.00 \\
         & Hits@100 & 0.03 & 0.24 & 0.00 & 0.04 & 0.03 \\
         & FHits@100 & 0.03 & 0.40 & 0.00 & 0.04 & 0.03 \\
         & macro\_MR & 490.14 & 1723.06 & 565.83 & 551.94 & 508.93 \\
         & micro\_MR & 513.98 & 2904.69 & 602.37 & 545.95 & 514.27 \\
         & macro\_FMR & 490.09 & 1642.98 & 565.80 & 551.89 & 508.92 \\
         & micro\_FMR & 513.97 & 2866.92 & 602.36 & 545.93 & 514.27 \\
         & macro\_AUC & 0.97 & 0.91 & 0.97 & 0.97 & 0.97 \\
         & micro\_AUC & 0.97 & 0.87 & 0.97 & 0.97 & 0.97 \\
         & macro\_FAUC & 0.97 & 0.92 & 0.97 & 0.97 & 0.97 \\
         & micro\_FAUC & 0.97 & 0.87 & 0.97 & 0.97 & 0.97 \\
\hline
\end{tabular}
\end{table}

\begin{table}[h!]
\centering
\begin{tabular}{| C{1.5cm} | C{2cm} | C{1.7cm} | C{1.7cm} | C{1.7cm} | C{1.7cm} | C{1.7cm} |}
\hline
& & ReLU & LeakyReLU & Reg & Losses & Neg. filter \\
\hline
Human hf & Hits@10 & 0.00 & 0.05 & 0.00 & 0.00 & 0.00 \\
         & FHits@10 & 0.00 & 0.06 & 0.00 & 0.00 & 0.00 \\
         & Hits@100 & 0.00 & 0.25 & 0.00 & 0.00 & 0.00 \\
         & FHits@100 & 0.00 & 0.26 & 0.00 & 0.00 & 0.00 \\
         & macro\_MR & 7642.19 & 5143.27 & 10295.57 & 3934.28 & 7617.16 \\
         & micro\_MR & 7645.75 & 4736.35 & 10226.14 & 3881.40 & 7608.48 \\
         & macro\_FMR & 7642.15 & 5132.24 & 10295.54 & 3934.24 & 7617.14 \\
         & micro\_FMR & 7645.73 & 4728.29 & 10226.12 & 3881.38 & 7608.47 \\
         & macro\_AUC & 0.85 & 0.90 & 0.79 & 0.92 & 0.85 \\
         & micro\_AUC & 0.85 & 0.92 & 0.80 & 0.93 & 0.85 \\
         & macro\_FAUC & 0.85 & 0.90 & 0.79 & 0.92 & 0.85 \\
         & micro\_FAUC & 0.85 & 0.92 & 0.80 & 0.93 & 0.85 \\
\hline
\end{tabular}
\end{table}

\section{ROC curves across different models}~\label{app:roc_curves}

First, we note that new functionality
incorporated into original {\it ELEmbeddings} model increases the
number of axioms localized within top-ranked subset while worsening
the ranking of axioms acquiring higher rank which enables more precise
model construction, especially for function prediction task. Although
ROC curve for PPI prediction in naive case illustrates that the
approach falls short compared to other models, it still outperforms
{\it ELEmbeddings} in terms of low-ranked axioms. Additional negative
losses carry out on algorithmic efficacy starting from higher-ranked
axioms, specially for yeast data. ROC curve for function prediction
tasks display the superior performance of naive predictor compared to
best learned geometric-based models.

\begin{figure}
\includegraphics[width=\textwidth]{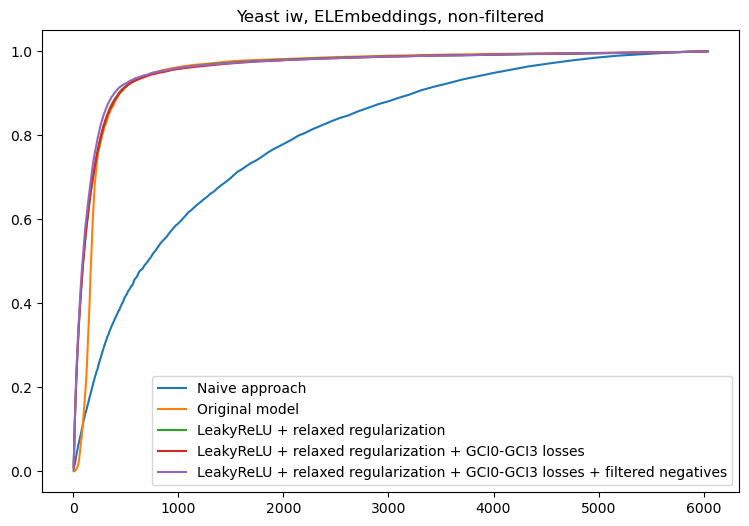}
\caption{ROC curves, Yeast iw dataset} \label{fig5}
\end{figure}

\FloatBarrier

\begin{figure}
\includegraphics[width=\textwidth]{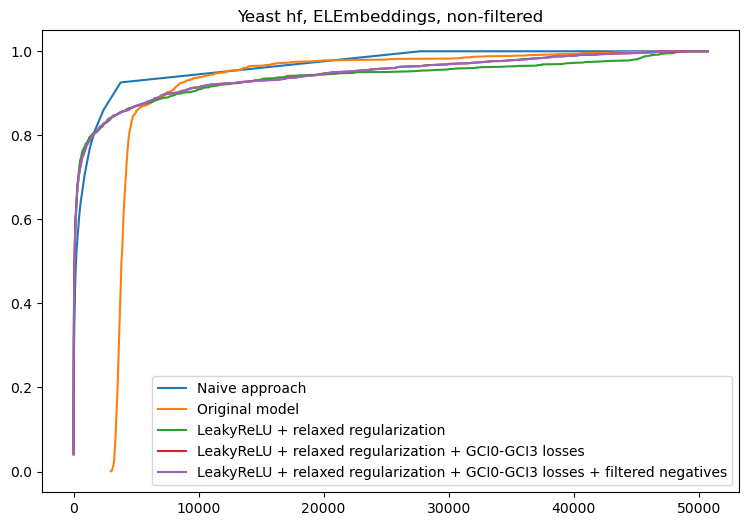}
\caption{ROC curves, Yeast hf dataset} \label{fig6}
\end{figure}

\begin{figure}
\includegraphics[width=\textwidth]{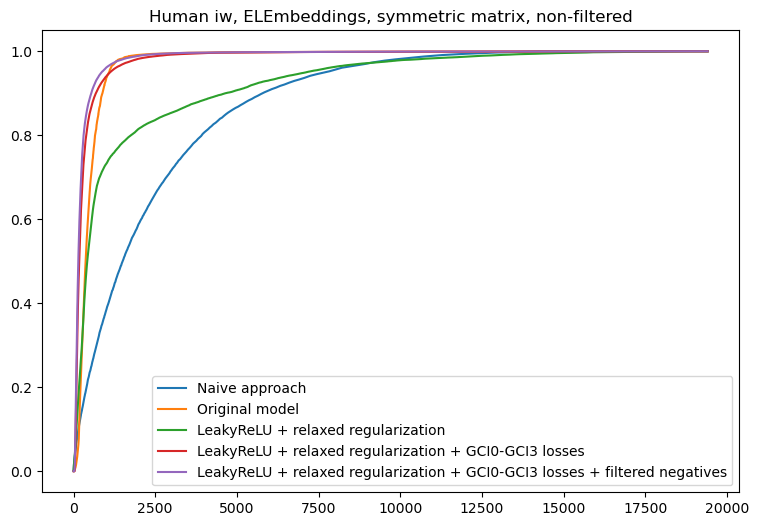}
\caption{ROC curves, Human iw dataset} \label{fig7}
\end{figure}

\begin{figure}
\includegraphics[width=\textwidth]{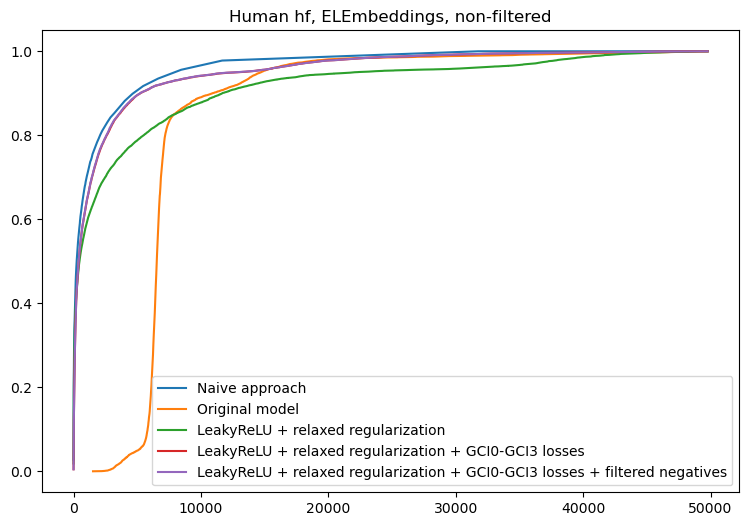}
\caption{ROC curves, Human hf dataset} \label{fig8}
\end{figure}

\section{ROC curves for LeakyReLU function}~\label{app:roc_curves_leaky_relu}

\begin{figure}[!h]
\includegraphics[width=\textwidth]{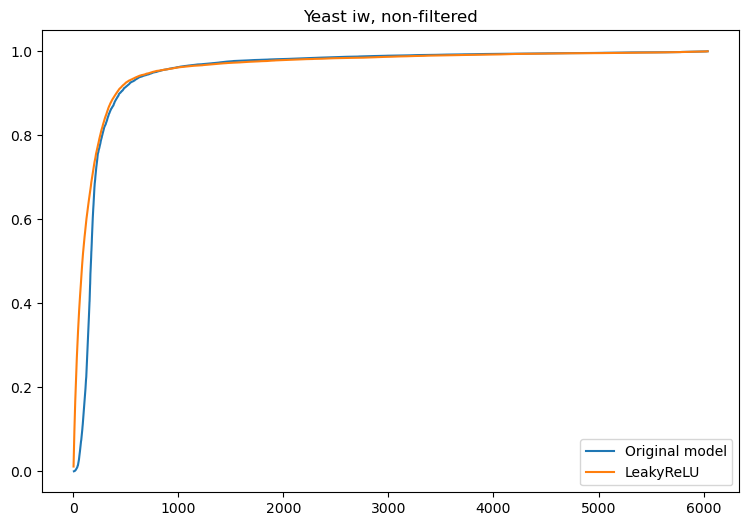}
\caption{ROC curves, Yeast iw dataset} \label{fig5}
\end{figure}

\begin{figure}
\includegraphics[width=\textwidth]{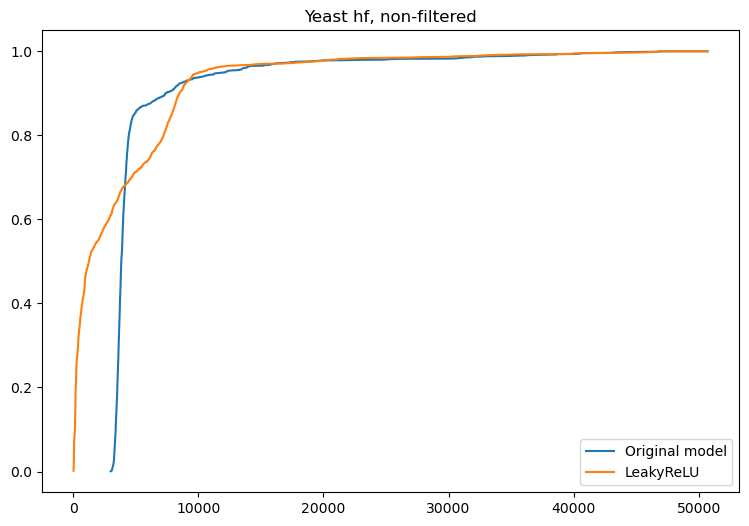}
\caption{ROC curves, Yeast hf dataset} \label{fig6}
\end{figure}

\begin{figure}
\includegraphics[width=\textwidth]{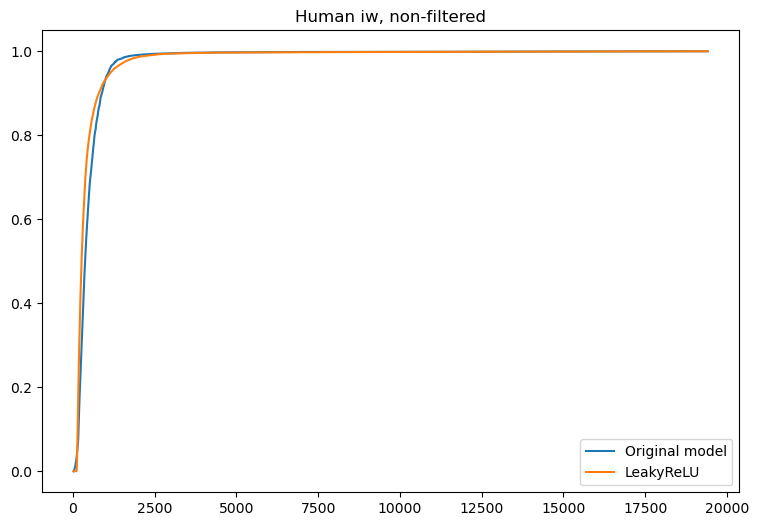}
\caption{ROC curves, Human iw dataset} \label{fig7}
\end{figure}

\begin{figure}
\includegraphics[width=\textwidth]{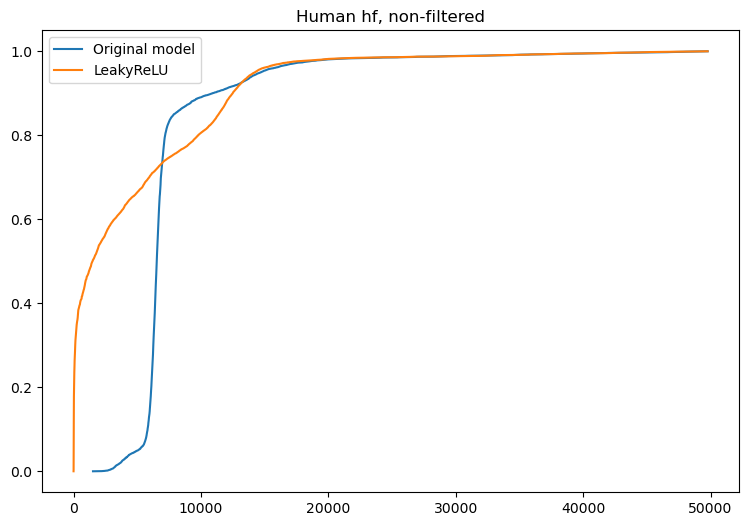}
\caption{ROC curves, Human hf dataset} \label{fig8}
\end{figure}

\section{Approximate deductive closure}~\label{sec:closure}

\begin{algorithm}[h!]\captionsetup{labelfont={sc,bf}, labelsep=newline} 
\caption{An algorithm for approximate computation of the deductive
  closure using inference rules; axioms in bold correspond to
  subclass/superclass axioms derived using ELK reasoner (here we use
  the transitive closure of the ELK inferences); plain axioms come
  from the knowledge base.}
\begin{algorithmic}
  \small
\For{all $C \sqsubseteq D$ in the knowledge base} \\
    \[
    \inference {C \sqsubseteq D \quad \boldsymbol{D \sqsubseteq D'}}{C \sqsubseteq D'} \quad 
    \inference {C \sqsubseteq D \quad \boldsymbol{C' \sqsubseteq C}}{C' \sqsubseteq D}
    \] \\
\EndFor
\For{all $C \sqcap D \sqsubseteq E$ in the knowledge base} \\
    \[
    \inference {C \sqcap D \sqsubseteq E \quad \boldsymbol{C' \sqsubseteq C}}{C' \sqcap D \sqsubseteq E}
    \] \\
\EndFor
\For{all $C \sqsubseteq \exists R.D$ in the knowledge base} \\
    \[
    \inference {C \sqsubseteq \exists R.D \quad \boldsymbol{D \sqsubseteq D'}}{C \sqsubseteq \exists R.D'} \quad 
    \inference {C \sqsubseteq \exists R.D \quad \boldsymbol{C' \sqsubseteq C}}{C' \sqsubseteq \exists R.D}
    \] \\
\EndFor
\For{all $\exists R.C \sqsubseteq D$ in the knowledge base} \\
    \[
    \inference {\exists R.C \sqsubseteq D \quad \boldsymbol{D \sqsubseteq D'}}{\exists R.C \sqsubseteq D'} \quad 
    \inference {\exists R.C \sqsubseteq D \quad \boldsymbol{C' \sqsubseteq C}}{\exists R.C' \sqsubseteq D}
    \] \\
\EndFor
\For{all $C \sqsubseteq \bot$ in the knowledge base} \\
    \[
    \inference {C \sqsubseteq \bot \quad \boldsymbol{C' \sqsubseteq C}}{C' \sqsubseteq \bot} 
    \]
\EndFor
\For{all $\exists R.C \sqsubseteq \bot$ in the knowledge base} \\
    \[
    \inference {\exists R.C \sqsubseteq \bot \quad \boldsymbol{C' \sqsubseteq C}}{\exists R.C' \sqsubseteq \bot}
    \]
\EndFor
\label{algo}
\end{algorithmic}
\end{algorithm}

\newpage

\section{Additional Figures}

\begin{figure}
\begin{subfigure}{0.49\textwidth}
     \includegraphics[width=\textwidth]{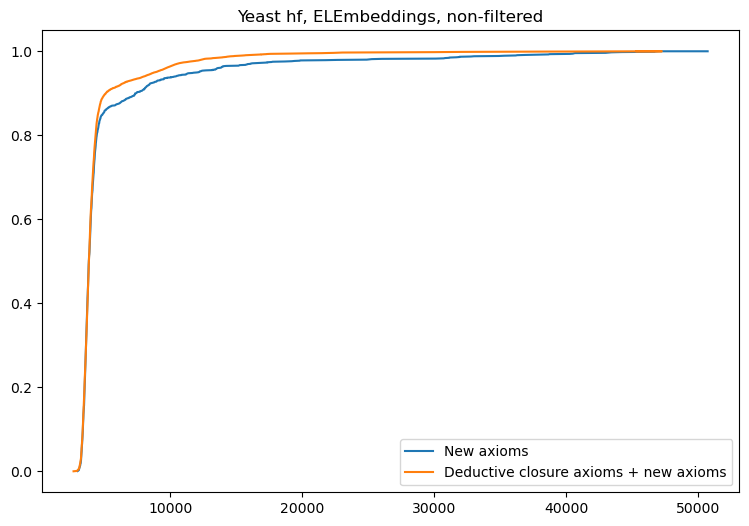}
     \label{fig:ab}
 \end{subfigure}
 \hfill
 \begin{subfigure}{0.49\textwidth}
     \includegraphics[width=\textwidth]{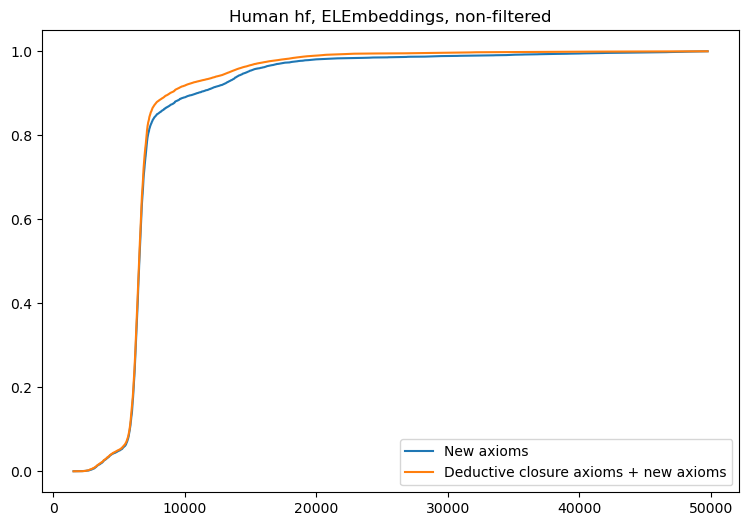}
     \label{fig:ba}
 \end{subfigure}
\caption{{\it ELEmbeddings}, ReLU, ROC curves for entailed axioms and novel axioms}
\label{fig:elem_dc}
\end{figure}

\begin{figure}
\begin{subfigure}{0.49\textwidth}
     \includegraphics[width=\textwidth]{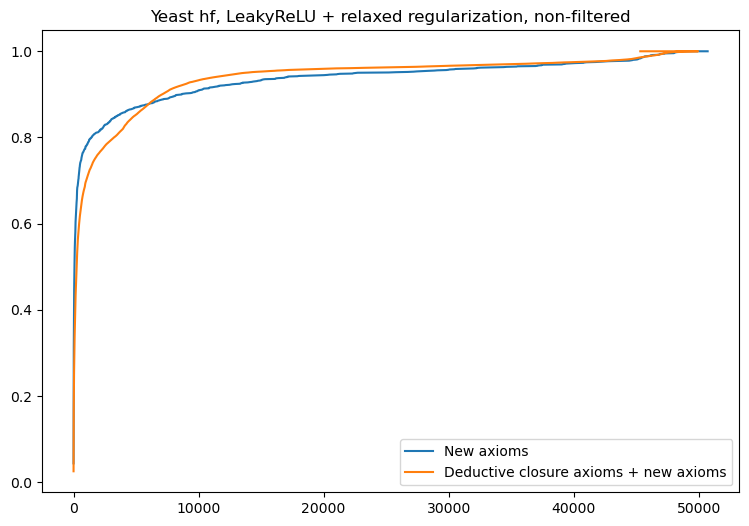}
     \label{fig:abb}
 \end{subfigure}
 \hfill
 \begin{subfigure}{0.49\textwidth}
     \includegraphics[width=\textwidth]{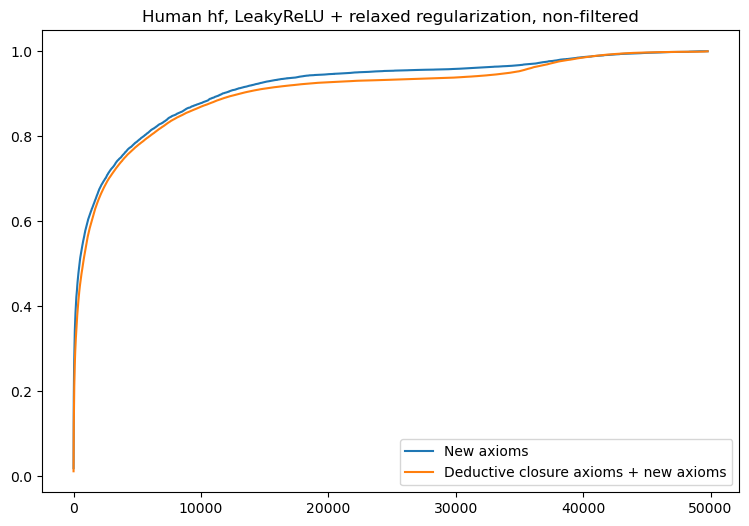}
     \label{fig:baa}
 \end{subfigure}
\caption{{\it ELEmbeddings}, LeakyReLU, ROC curves for entailed axioms and novel axioms}
\label{fig:elem_leaky_dc}
\end{figure}

\begin{figure}
\begin{subfigure}{0.49\textwidth}
     \includegraphics[width=\textwidth]{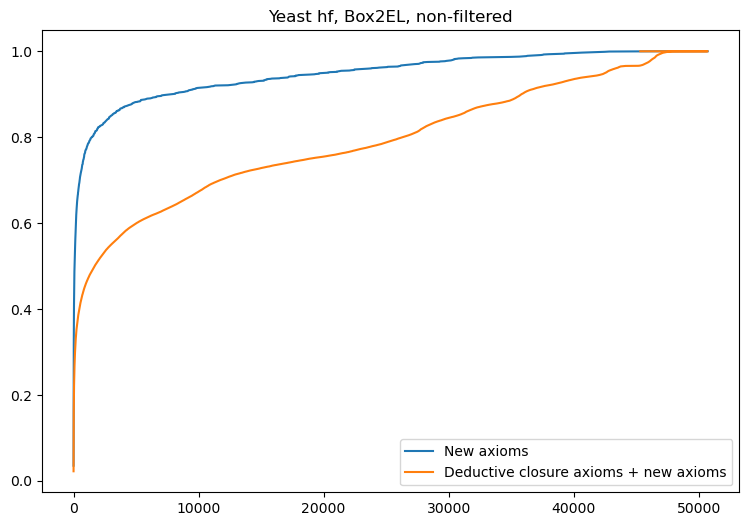}
     \label{fig:abbb}
 \end{subfigure}
 \hfill
 \begin{subfigure}{0.49\textwidth}
     \includegraphics[width=\textwidth]{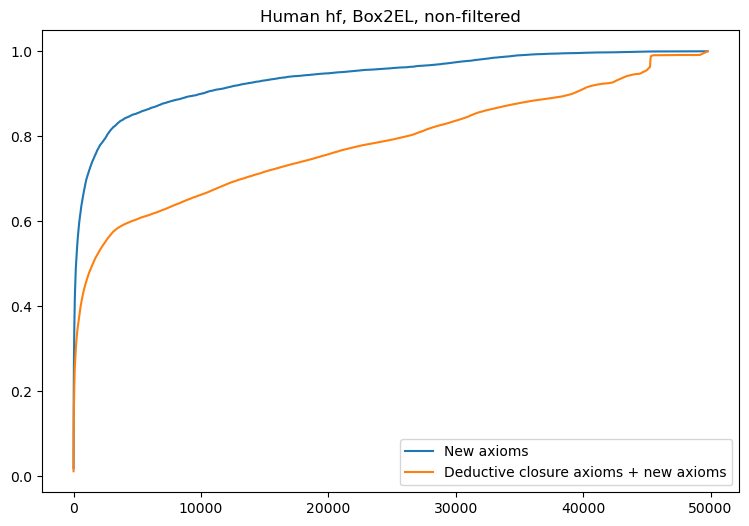}
     \label{fig:baaa}
 \end{subfigure}
\caption{{\it $Box^2EL$}, ROC curves for entailed axioms and novel axioms}
\label{fig:box2el_dc}
\end{figure}

\end{document}